\documentclass[10pt,twocolumn,letterpaper]{article}

\usepackage[pagenumbers]{cvpr} 

\usepackage{graphicx}
\usepackage{amsmath}
\usepackage{amssymb}
\usepackage{booktabs}

\usepackage{algorithm}
\usepackage{algpseudocode}
\usepackage{enumitem}
\usepackage{mathtools}
\usepackage{makecell}
\usepackage{comment}
\usepackage{pifont}
\usepackage[dvipsnames]{xcolor}
\usepackage{anyfontsize}

\usepackage[accsupp]{axessibility}
\usepackage[title,titletoc,page]{appendix}

\usepackage[pagebackref,breaklinks,colorlinks]{hyperref}

\usepackage[capitalize]{cleveref}
\crefname{section}{Sec.}{Secs.}
\Crefname{section}{Section}{Sections}
\Crefname{table}{Table}{Tables}
\crefname{table}{Tab.}{Tabs.}

\def\cvprPaperID{1223} 
\def\confName{CVPR}
\def\confYear{2023}

\begin{document}

\title{RGB no more: Minimally-decoded JPEG Vision Transformers}
\author{Jeongsoo Park \qquad\qquad Justin Johnson\\
University of Michigan\\
{\tt \small \{jespark, justincj\}@umich.edu}
}
\maketitle

\begin{abstract}
    \urlstyle{rm}
    Most neural networks for computer vision are designed to infer using RGB images. However, these RGB images are commonly encoded in JPEG before saving to disk; decoding them imposes an unavoidable overhead for RGB networks. Instead, our work focuses on training Vision Transformers (ViT) directly from the encoded features of JPEG. This way, we can avoid most of the decoding overhead, accelerating data load. Existing works have studied this aspect but they focus on CNNs. Due to how these encoded features are structured, CNNs require heavy modification to their architecture to accept such data. Here, we show that this is not the case for ViTs. In addition, we tackle data augmentation directly on these encoded features, which to our knowledge, has not been explored in-depth for training in this setting. With these two improvements -- ViT and data augmentation -- we show that our ViT-Ti model achieves up to 39.2\% faster training and 17.9\% faster inference with no accuracy loss compared to the RGB counterpart. \footnote{Code available at \footnotesize \fontsize{8}{10}{\href{https://github.com/JeongsooP/RGB-no-more}{https://github.com/JeongsooP/RGB-no-more}}}
\end{abstract}

\setlength{\textfloatsep}{10pt}
\setlength{\dbltextfloatsep}{10pt}
\setlength{\dblfloatsep}{5pt}
\setlength{\intextsep}{3pt}
\setlength{\floatsep}{5pt}
\setlength{\abovedisplayskip}{3pt}
\setlength{\belowdisplayskip}{3pt}

\vspace{-0.53cm}
\section{Introduction}
\label{sec:intro}

Neural networks that process images typically receive their inputs as regular grids of RGB pixel values.
This \emph{spatial-domain} representation is intuitive, and matches the way that images are displayed on digital devices (e.g. LCD panels with RGB sub-pixels).
However, images are often stored on disk as compressed JPEG files that instead use \emph{frequency-domain} representations for images.
In this paper we design neural networks that can directly process images encoded in the frequency domain.

\begin{figure}[t]
    \centering
    \includegraphics[width=\columnwidth]{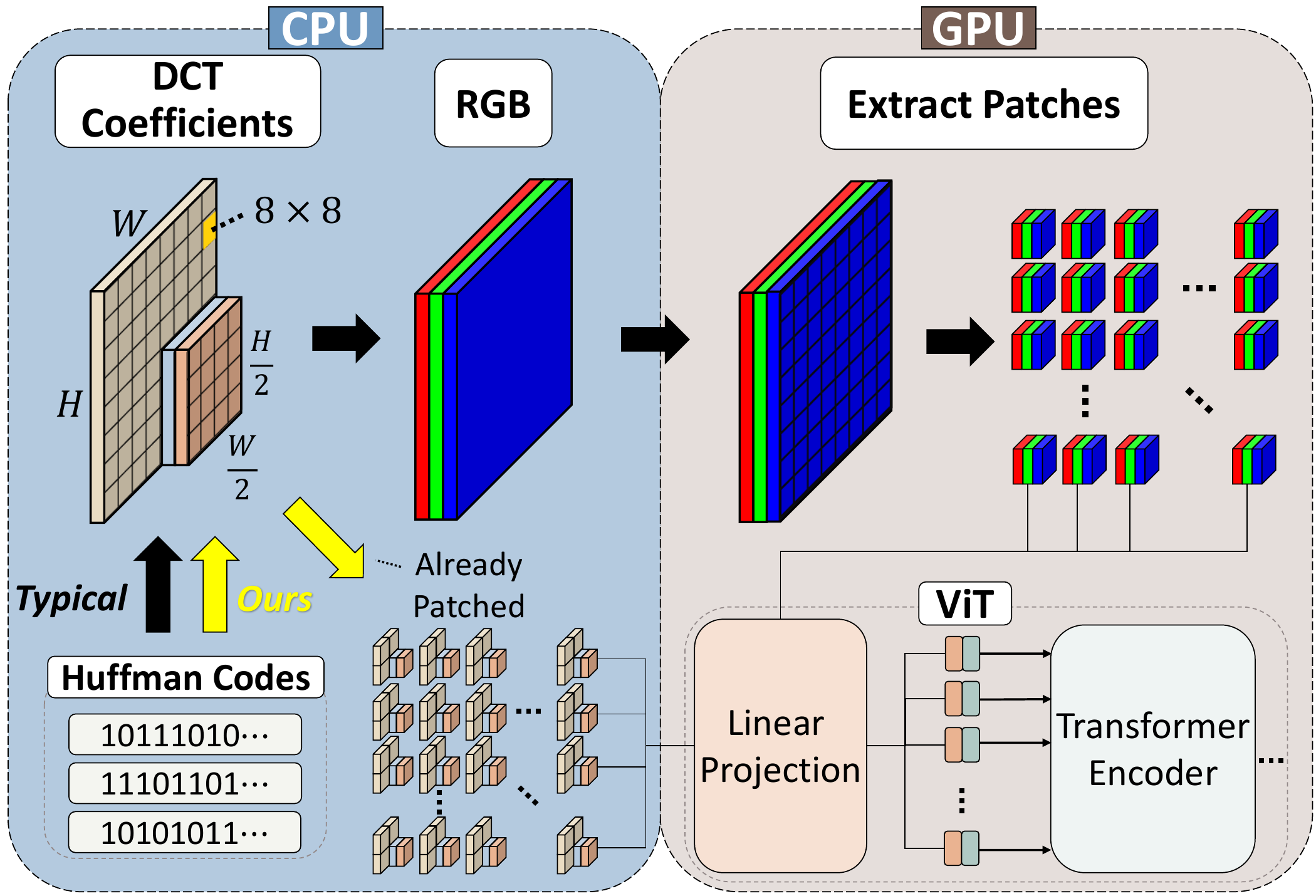}
    \caption{Our proposed training process. The typical process requires full decoding as well as patch extraction to train. In contrast, our process does not since the DCT coefficients are already saved in block-wise fashion. As ViTs work on image patches, we can directly feed these coefficients to the network.}
    \label{fig:training_processes}
\end{figure}

Networks that process frequency-domain images have the potential for much faster data loading.
JPEG files store image data using Huffman codes; these are decoded to (frequency-domain) discrete cosine transform (DCT) coefficients then converted to (spatial-domain) RGB pixels before being fed to the neural network (\cref{fig:training_processes}).
Networks that process DCT coefficients can avoid the expensive DCT to RGB conversion;
we show in \cref{bak:jpegcomp_cost} that this can reduce the theoretical cost of data loading by up to 87.6\%.
Data is typically loaded by the CPU while the network runs on a GPU or other accelerator;
more efficient data loading can thus reduce CPU bottlenecks and accelerate the entire pipeline.

We are not the first to design networks that process frequency-domain images.
The work of Gueguen \etal\cite{fasternnjpeg} and Xu \etal \cite{learninginfreq} are most similar to ours: they show how standard CNN architectures such as ResNet~\cite{he2016deep} and MobileNetV2~\cite{sandler2018mobilenetv2}
can be modified to input DCT rather than RGB and trained to accuracies comparable to their standard formulations.
We improve upon these pioneering efforts in two key ways: \emph{architecture} and \emph{data augmentation}.

Adapting a CNN architecture designed for RGB inputs to instead receive DCT is nontrivial.
The DCT representation of an $H\times W\times 3$ RGB image consists of 
a $\frac{H}{8}\times\frac{W}{8}\times8\times 8$ tensor of \emph{luma} data and two $\frac{H}{16}\!\times\frac{W}{16}\times 8 \times 8$ tensors of \emph{chroma} data.
The CNN architecture must be modified both to accept lower-resolution inputs (e.g. by skipping the first few stages of a ResNet50 and adding capacity to later stages) and to accept heterogeneously-sized luma and chroma data (e.g. by encoding them with separate pathways).

We overcome these challenges by using Vision Transformers (ViTs) \cite{vitpaper} rather than CNNs.
ViTs use a \emph{patch embedding} layer to encode non-overlapping image patches into vectors, which are processed using a Transformer~\cite{vaswani2017attention}.
This is a perfect match to DCT representations, which also represent non-overlapping RGB image patches as vectors.
We show that ViTs can be easily adapted to DCT inputs by modifying only the initial patch embedding layer and leaving the rest of the architecture unchanged.

Data augmentation is critical for training accurate networks; this is especially true for ViTs~\cite{deit, howtotrainvit, vit_datahungry}.
However, standard image augmentations such as resizing, cropping, flipping, color jittering, \etc are expressed as transformations on RGB images; prior work \cite{fasternnjpeg,learninginfreq} on neural networks with DCT inputs thus implement data augmentation by converting DCT to RGB, augmenting in RGB, then converting back to DCT before passing the image to the network.
This negates all of the potential training-time efficiency gains of using DCT representations;
improvements can only be realized during inference when augmentations are not used.

We overcome this limitation by augmenting DCT image representations directly, avoiding any DCT to RGB conversions during training.
We show how all image augmentations used by RandAugment~\cite{randaugment} can be implemented on DCT representations.
Some standard augmentations such as image rotation and shearing are costly to implement in DCT, so we also introduce several new augmentations which are natural for DCT.

Using these insights, we train ViT-S and ViT-Ti models on ImageNet~\cite{imagenet,russakovsky2015imagenet} which match the accuracy of their RGB counterparts.
Compared to an RGB equivalent, our ViT-Ti model is up to 39.2\% faster per training iteration and 17.9\% faster during inference.
We believe that these results demonstrate the benefits of neural networks that ingest frequency-domain image representations.

\newcommand{\ourpar}[1]{\vspace{2pt}\noindent\textbf{#1}}

\section{Related Work}
\label{sec:related}
\ourpar{Training in the frequency domain} is extensively explored in the recent studies. 
They consider JPEG \cite{fasternnjpeg,learninginfreq,imgcomplearn1_j2k,imgcomplearn3_fpga,imgcomplearn5_CNN_cifar,imgcomplearn6_CNN,imgcomplearn7_CNN_uberlike,imgcomplearn8_CNN_uberlike2,imgcomplearn9_CNN,imgcomplearn13_deepres,Yousfi2020AnIS}, DCT \cite{imgcomplearn2_lane,imgcomplearn4_resnettumor,imgcomplearn10_CNN,imgcomplearn11_AEnc,imgcomplearn12_tinyCNN,imgcomplearn14_objdetect,imgcomplearn15_objdetect2, Goldberg2020RethinkingFF} or video codecs \cite{videocomplearn2_hevc,videocomplearn3_actrecog,compvidactionrecog,videocomplearn4} with a primary focus on increasing the throughput of the model by skipping most of the decoding steps. Many of these works base their model architecture on CNNs. However, adapting CNNs to accept frequency input requires nontrivial modification to the architecture \cite{fasternnjpeg,learninginfreq,imgcomplearn7_CNN_uberlike,imgcomplearn8_CNN_uberlike2,videocomplearn2_hevc}. More recent studies \cite{compressedvision,videocomplearn1_neuralcompressor} explore training from a neural compressor \cite{neuralcomp1,neuralcomp2,neuralcomp3,neuralcomp4,neuralcomp5,neuralcomp6,neuralcomp7,neuralcomp8,neuralcomp9,neuralcomp10} instead of an existing compression algorithms \cite{jpegformat,jpegjfifrec,HEVCstandard,HEVCandVidCodecs}. This approach, however, requires transcoding the existing data to their neural compressed format, increasing overhead. We instead use Vision Transformers \cite{vitpaper,deit,howtotrainvit,betterbaselinevit} on the JPEG-encoded data. Our approach has two advantages: (1) patch-wise architecture of ViTs is better suited for existing compression algorithms, (2) does not require any transcoding; it can work on any JPEG images.

\ourpar{Data augmentation directly on the frequency domain} 
has been studied in several works.
Gueguen \etal \cite{fasternnjpeg} suggested augmenting on RGB and converting back to DCT. Wiles \etal \cite{compressedvision} used an augmentation network that is tailored towards their neural compressor. Others focus on a more classical approach such as sharpening 
\cite{jpegsharpen1,dctblurdetection1,dctblurdetection2,dctblurdetection3,jpegsharpen2,jpegsharpen3,jpegsharpen4}, resizing \cite{jpegresize1,jpegresize2_coding,jpegresize3_fast,jpegresize4_lmfold,jpegresize5_scalable,subband,subblock,dctresizing}, watermarking \cite{jpegwatermark1_hidden,jpegwatermark2,jpegwatermark3,jpegwatermark4,jpegwatermark5,jpegwatermark6,jpegwatermark7}, segmentation \cite{DCTsegment5,DCTsegment2,DCTsegment3,DCTsegment4,DCTsegment1}, flip and 90-degree rotation \cite{jpegfliprot90,Yousfi2020AnIS}, scalar operations \cite{jpegscalar}, and forgery detection \cite{jpegforgery1,jpegforgery2,jpegforgery3_resample} via analyzing the properties of JPEG and DCT. However, to our knowledge, no other works have thoroughly studied the effect of DCT augmentation during frequency-domain training.

\label{rel:fastervit}
\ourpar{Speeding up ViTs} has been rigorously studied. These either utilize CNN-hybrid architectures to reduce computation
\cite{vit1_mobilevit_cnnhybrid,vit3_edgenext_cnnhybrid,vit6_mocovit_cnnhybrid,vit12_lightvit_cnnhybrid,vit13_lightvit_cnnhybrid,vit14_fastvit_cnnhybrid,vit19_light_cnnhybrid}, speed up attention by sparsifying \cite{vit2_edgevit_purevit,vit8_lightvit_purevit,vit9_fastervit_purevit,vit17_swin_faster_purevit}, linearizing \cite{vit5_efficientvit_purevit,vit4_purevit}, or through other techniques such as pruning
\cite{vit10_fastervit_purevit_pruning,vit7_efficientformer_purevit,vit11_accvit_purevit_pruning,vit18_pruning_purevit}, bottlenecking \cite{vit15_fastatt_purevit}, or approximation \cite{vit16_obj_purevit}. While we only consider the plain ViT architecture in this paper, we want to emphasize that faster data loading is imperative to fully take advantage of these speed-ups as models can only infer as fast as the data loading speed.

\section{Background}
\label{sec:background}

\numberwithin{equation}{section}

Here, we discuss Discrete Cosine Transform (DCT) and JPEG compression process. They are crucial to designing a model and DCT data augmentations in the later sections.

\begin{figure}[t]
    \centering
    \includegraphics[width=\columnwidth]{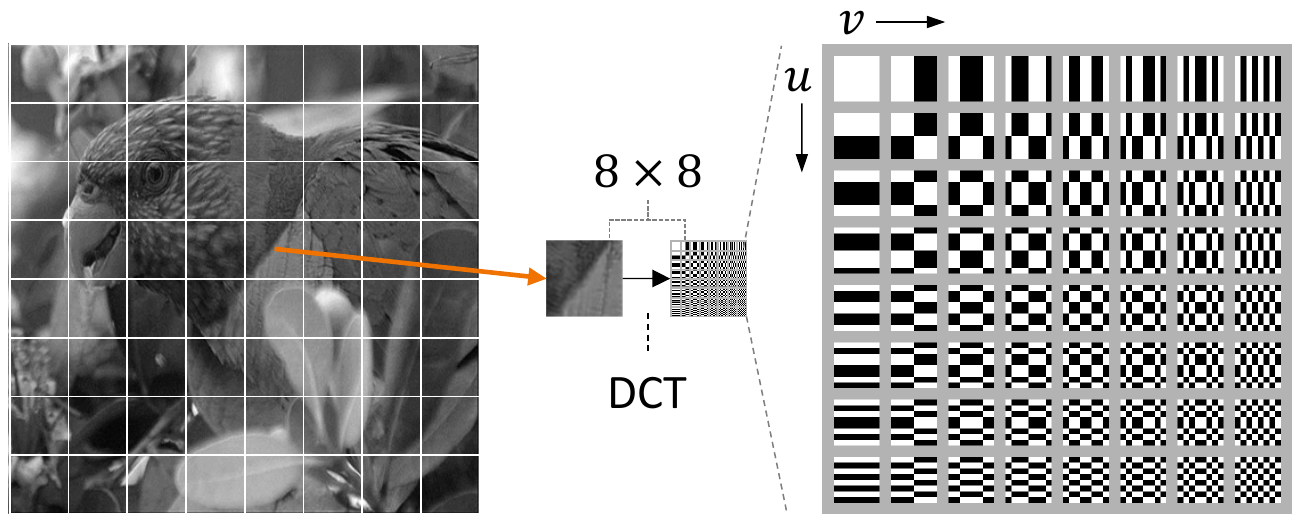}
    \caption{Process of applying $8 \times 8$ DCT to the input image. The input image is sliced into $8 \times 8$ patches and the DCT is applied to each patch. The DCT bases are shown on the right.}
    \label{fig:dctbases}
\end{figure}

\label{bak:DCT}
\ourpar{Discrete cosine transform} decomposes a finite data sequence into a sum of discrete-frequency cosine functions. It is a transformation from the spatial domain to the frequency domain. We will focus on $8 \times 8$ DCT since it is a transform used in JPEG.
Let $x \in \mathbb{R}^{8 \times 8}$ be a $8 \times 8$ image patch. Then its DCT transform $X \in \mathbb{R}^{8 \times 8}$ is given by:
\begin{equation}
\resizebox{\columnwidth}{!}{$
\displaystyle X_{u,v} = \frac{\alpha_u \alpha_v}{4} \sum_{m,n} x_{m,n} \cos\big[\frac{\pi(2m+1)u}{16}\big]\cos\big[\frac{\pi(2n+1)v}{16}\big]
$}
\label{eq:dct8by8}
\end{equation}
Where $\alpha_i = 1/\sqrt{2}$ if $i=0$, else $1$, $u,v,m,n \in [0..7]$.
\Cref{fig:dctbases} shows how the DCT is applied to an image in JPEG. The original image patch can be reconstructed by a weighted sum of the DCT bases (\cref{fig:dctbases}) and their corresponding coefficients $X_{u,v}$.
\label{bak:DCTscale}
For the standard JPEG setting, the pixel-space min/max value of $[-128, 127]$, is scaled up by $8\times$ to $[-1024, 1016]$. 
The proof of this property is shown in \cref{apdx:scalingDCT}. This property is necessary to implement several DCT augmentations in \cref{sec:DCTAug}.

\begin{figure}[t]
    \centering
    \includegraphics[width=\columnwidth]{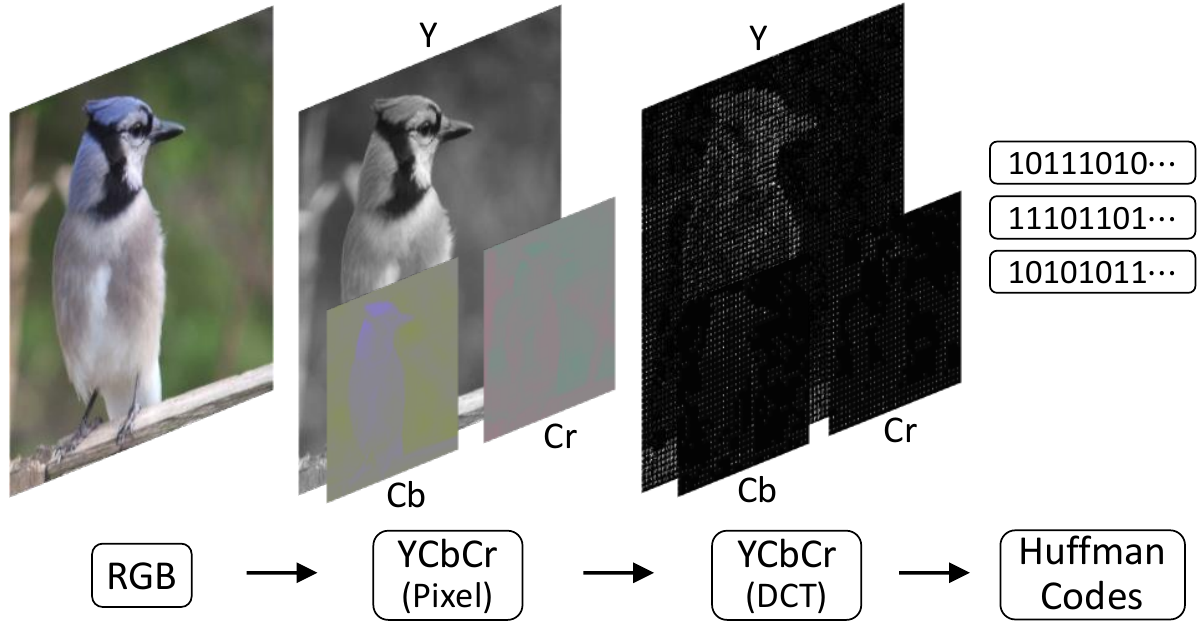}
    \caption{A simplified JPEG compression process. An RGB image is first converted to YCbCr, then transformed to DCT space. They are then encoded into binary codes and written to disk. Decoding follows the inverse of this process.}
    \label{fig:jpegproc}
\end{figure}

\label{bak:jpegcomp}
\ourpar{JPEG} \cite{jpegformat,jpeg25years,jpegjfifrec} is a
widely used compression algorithm that is designed to encode images generated by digital photography. 
The encoding process is as follows:
\begin{enumerate}[label=(\alph*)]
    \itemsep-3pt
    \item {\small$H\times W\times 3$} RGB image is given as input 
    \item RGB is converted to YCbCr color space 
    \item CbCr channels are downsampled to $\frac{H}{2}\times\frac{W}{2}$ \label{enum:jpegsubsample} 
    \item Values are shifted from [0,255] to [-128,127] \label{enum:zerocentered} 
    \item DCT is applied to non-overlapping $8\times8$ pixel patches\label{enum:dctpatch} 
    \item DCT coefficients are quantized \label{enum:jpegquant} 
    \item Run-length encoding{\small(RLE)} compresses the coefficients  
    \item RLE symbols are encoded using Huffman coding 
\end{enumerate} 
A simplified illustration of this process is shown in \Cref{fig:jpegproc}. YCbCr is a color space where Y represents luma (i.e. brightness) and Cb, Cr signifies chroma (i.e. color) of the image.
Step \labelcref{enum:dctpatch} produces a data of size $\mathbf{Y} \in \mathbb{R}^{1\times \frac{H}{8} \times \frac{W}{8} \times 8 \times 8}, \mathbf{U} \in \mathbb{R}^{2 \times \frac{H}{16} \times \frac{W}{16} \times 8 \times 8}$ for Y and CbCr channel respectively. These $8 \times 8$ DCT coefficients are referred to as \textit{DCT Blocks} in the later sections.

\label{bak:jpegcomp_cost}
\ourpar{Compute cost to decode JPEG} can be analyzed by counting the number of operations (OPs) for the inverse of the above process. 
Consider decoding a single $8 \times 8$ patch. Our proposed scheme decodes step (h) - (f) with a computation cost of $3N_s + 128$ OPs where $N_s \in [1..64]$: number of RLE symbols. Full JPEG decoding, on the other hand, requires $3N_s + 1717$ OPs. If we suppose $N_s = 32$, then the compute cost is $224$ and $1813$ OPs respectively, where our scheme theoretically saves computation by $87.6\%$. The details of these values are shown in \cref{apdx:jpegcost_detail}.

\section{Model Architecture}
\label{sec:ModelArch}

\begin{figure*}
    \centering
    \includegraphics[width=0.9\textwidth]{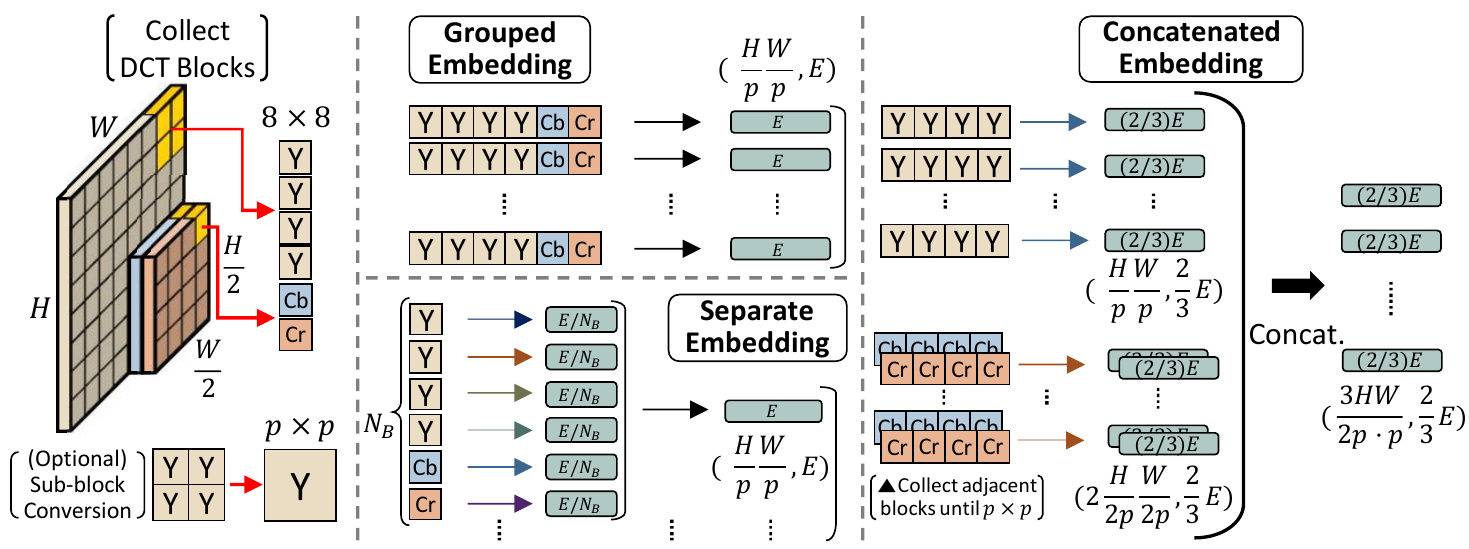}
    \caption{Proposed embedding strategies. We first collect $8 \times 8$ DCT blocks until it matches the patch size. Then, these blocks can be embedded using one of the strategies. Grouped embedding groups all of the collected blocks and embeds them together. Separate embedding embeds each block separately and mixes it. Concatenated embedding embeds $\mathbf{Y}$ and $\mathbf{U}$ separately and concatenates them.}
    \label{fig:model_architectures}
\end{figure*}
Designing a neural network that works on JPEG-encoded DCT coefficients can be challenging. In \cref{bak:jpegcomp}, we showed that JPEG downsamples CbCr channels to $\frac{H}{2} \times \frac{W}{2}$. This spatial disparity must be addressed before training in DCT. An existing work by Gueguen \etal \cite{fasternnjpeg} suggested several new CNN architectures which include 
(1) upsampling, (2) downsampling, and (3) late-concatenation. The first two architectures either upsample CbCr to match the dimension of Y or downsample Y to match CbCr. However, doing so results in (1) redundant computation or (2) loss of information due to resizing. The third approach, late-concatenation, computes them separately and concatenates them further down the network. However, this requires substantial modification of the CNN architecture, making adaptation to existing models difficult.

We believe that ViTs are better suited to deal with this unique characteristic of JPEG. Vision transformers work on patches of an image \cite{vitpaper,deit,howtotrainvit,vit_datahungry}. Considering that JPEG DCT already extracts $8 \times 8$ patches from an image (\cref{bak:jpegcomp} \labelcref{enum:dctpatch}), we can employ them by modifying only the initial embedding layer. 
This allows easier integration into other ViTs as the rest of the architecture can remain untouched.

Therefore, in this section, we propose several patch-embedding strategies that are plausible solutions to this problem. These modified patch embedding layers are illustrated in \cref{fig:model_architectures}. The architecture that follows is identical to the plain ViT defined in \cite{vitpaper,betterbaselinevit,deit}.

\ourpar{Grouped embedding} generates embeddings by grouping the $8 \times 8$ DCT blocks together from the corresponding patch position. 
Consider a DCT input $\mathbf{Y}, \mathbf{U}$ defined in \cref{bak:jpegcomp}. Grouped embedding collects DCT blocks such that 
for patch size $p$ \footnote{Assume $p = 16n, n\in\mathbb{N}$. For $p<16$, check \cref{apdx:smallerpatch}.},
$\mathbf{Y} \rightarrow \mathbf{Y}_r \in \mathbb{R}^{\frac{H}{p} \times \frac{W}{p} \times p^2}$ and $\mathbf{U} \rightarrow \mathbf{U}_r \in \mathbb{R}^{\frac{H}{p} \times \frac{W}{p} \times \frac{2p^2}{4}}$ where the channel and block size are flattened to the last dimension.
Then, this is concatenated along the last axis as $(\mathbf{Y}_r,\mathbf{U}_r) \rightarrow \mathbf{YU}_r \in \mathbb{R}^{\frac{H}{p} \times \frac{W}{p} \times \frac{3p^2}{2}}$ which will then be embedded as $\mathbf{z}: \mathbf{YU}_r \rightarrow \mathbf{z} \in \mathbb{R}^{\frac{HW}{p\cdot p} \times E}$ where $\mathbf{z}$ is the generated embedding and $E$ is the embedding size.

\ourpar{Separate embedding} generates separate embeddings for each DCT block in a patch. A DCT input $\mathbf{Y}, \mathbf{U}$ is reshaped as $\mathbf{Y} \rightarrow \mathbf{Y}_r \in \mathbb{R}^{\frac{H}{p} \times \frac{W}{p} \times \frac{p^2}{64} \times 64}, \mathbf{U} \rightarrow \mathbf{U}_r \in \mathbb{R}^{\frac{H}{p} \times \frac{W}{p} \times \frac{2p^2}{4\cdot 64} \times 64}$ which is embedded separately for each block: $(\mathbf{Y}_r, \mathbf{U}_r) \rightarrow \mathbf{z} \in \mathbb{R}^{\frac{HW}{p\cdot p} \times \frac{3p^2}{2\cdot64} \times \frac{E}{N_B}}$ where $N_B$ = number of blocks = $\frac{3p^2}{2 \cdot 64}$. This is then mixed using a linear layer to generate a final embedding $\mathbf{z} \rightarrow \mathbf{z} \in \mathbb{R}^{\frac{HW}{p\cdot p} \times E}$.
Our intuition behind this strategy is that the information each block holds might be critical, thus training a specialized linear layer for each block could yield better results.

\ourpar{Concatenated embedding} embeds and concatenates the DCT blocks from $\mathbf{Y}$, $\mathbf{U}$ separately. Our intuition is that since $\mathbf{Y}$ and $\mathbf{U}$ represent different information (luma and chroma), designing specialized layers that handle each information separately may be necessary. However, this generates more embeddings per image patch than the plain model.
To keep the overall size even, we reduce the size of each embedding to 2/3. An embedding formula is $\mathbf{Y} \rightarrow \mathbf{Y}_r \in \mathbb{R}^{\frac{H}{p} \times \frac{W}{p} \times p^2}, \mathbf{U} \rightarrow \mathbf{U}_r \in \mathbb{R}^{2 \times \frac{H}{2p} \times \frac{W}{2p} \times p^2}$, which is then embedded separately per channel type: $\mathbf{Y}_r \rightarrow \mathbf{z}_Y \in \mathbb{R}^{\frac{HW}{p\cdot p} \times \frac{2E}{3}}, \mathbf{U}_r \rightarrow \mathbf{z}_U \in \mathbb{R}^{\frac{2HW}{4p\cdot p} \times \frac{2E}{3}}$ then concatenated $(\mathbf{z}_Y,\mathbf{z}_U) \rightarrow \mathbf{z} \in \mathbb{R}^{\frac{3HW}{2p \cdot p} \times \frac{2E}{3}}$ to generate an embedding $\mathbf{z}$.

\ourpar{Sub-block conversion} \cite{subblock} can be applied as an alternate way to embed a patch. Consider ViT architecture of patch size 16. For simplicity, assume only the Y channel is present. To form a patch size of $16 \times 16$, four $8 \times 8$ DCT blocks have to be grouped together. One strategy is to embed these directly through the linear layer. Another approach is to convert them into a single $16 \times 16$ block and embed them. In other words, we embed the DCT patches from a $16 \times 16$ DCT. There exists a way to efficiently extract these $16 \times 16$ DCT from the smaller $8 \times 8$ DCT blocks and vice versa known as \textit{sub-block conversion} \cite{subblock}. This technique can allow us to extract a native DCT patch of different sizes, potentially yielding better results. We also use this technique to implement several augmentations in \cref{sec:DCTAug}.

\section{DCT Augmentation}
\label{sec:DCTAug}

Data augmentation has been a vital component in training robust networks \cite{randaugment,dataaug1,dataaug2,dataaug3,dataaug4,dataaug5,dataaug6-autoaug}. However, augmenting the DCT, as well as training with it, has not been studied in depth. There exist several prior works for some augmentations such as sharpening or resizing as discussed in \cref{sec:related}, but most other RGB augments lack their DCT counterparts.

Existing work by Gueguen \etal \cite{fasternnjpeg} proposed converting DCT to RGB, augmenting in RGB, and converting it back to DCT. However, this incurs expensive RGB/DCT conversion as shown in \cref{bak:jpegcomp_cost}. More recent work by Wiles \etal \cite{compressedvision} used a specialized augmentation network that augments their neural-compressed format. Doing so, however, sacrifices versatility as it requires training and can't be reliably generalized to other resolutions or data.

Our approach is different. We instead implement augmentations directly on DCT by analyzing its properties. That way, we can avoid converting to RGB or relying on a trained augmentation network. 
In other words, our method is fast, flexible, and works on virtually any data, so long as it is represented in DCT. Thus, in this section, we implement all augmentations used in RandAugment \cite{randaugment} as well as suggest new augmentations that are meaningful for DCT. We mark the ones we suggest using an asterisk(*).

\begin{figure*}[t]
    \centering
    \includegraphics[width=\textwidth]{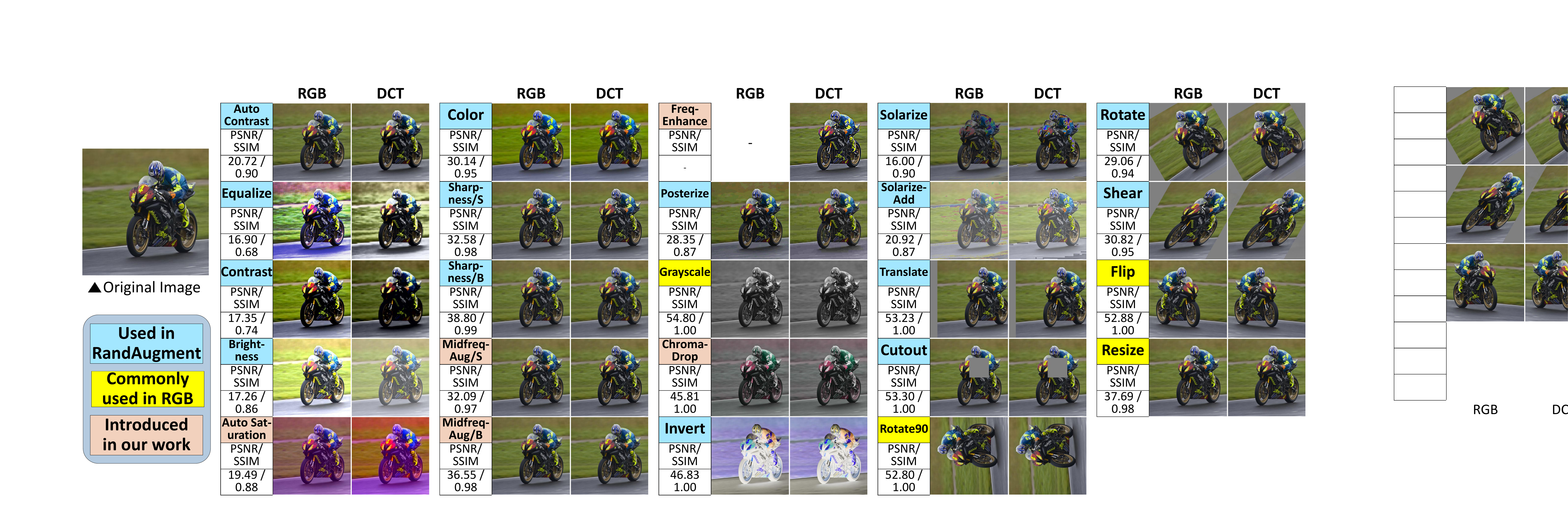}
    \caption{A visualization of RGB augmentations and their DCT counterparts. `/S' or `/B' indicates `sharpen' or `blur' on sharpness-related augmentations. Resize has been compared where they scale the image up by 2. PSNR and SSIM are calculated for each DCT augmentation.
    We observe that most DCT augmentations resemble RGB augmentations. Augmentations such as grayscale, cutout, and flip are  identical to RGB. However, other augmentations including equalize and contrast are not exactly equal to RGB.} 
    \label{fig:randaugments}
\end{figure*}

There are largely two different types of DCT augmentation: photometric and geometric. Each of which uses different key properties of the DCT. While these augmentations are not meant to precisely reproduce RGB augmentations, most of our implementations approximate the RGB counterparts reasonably well. The overview of the augmentations and their similarity metrics are shown in \cref{fig:randaugments}.

\begin{figure}[t]
    \centering
    \includegraphics[width=0.85\columnwidth]{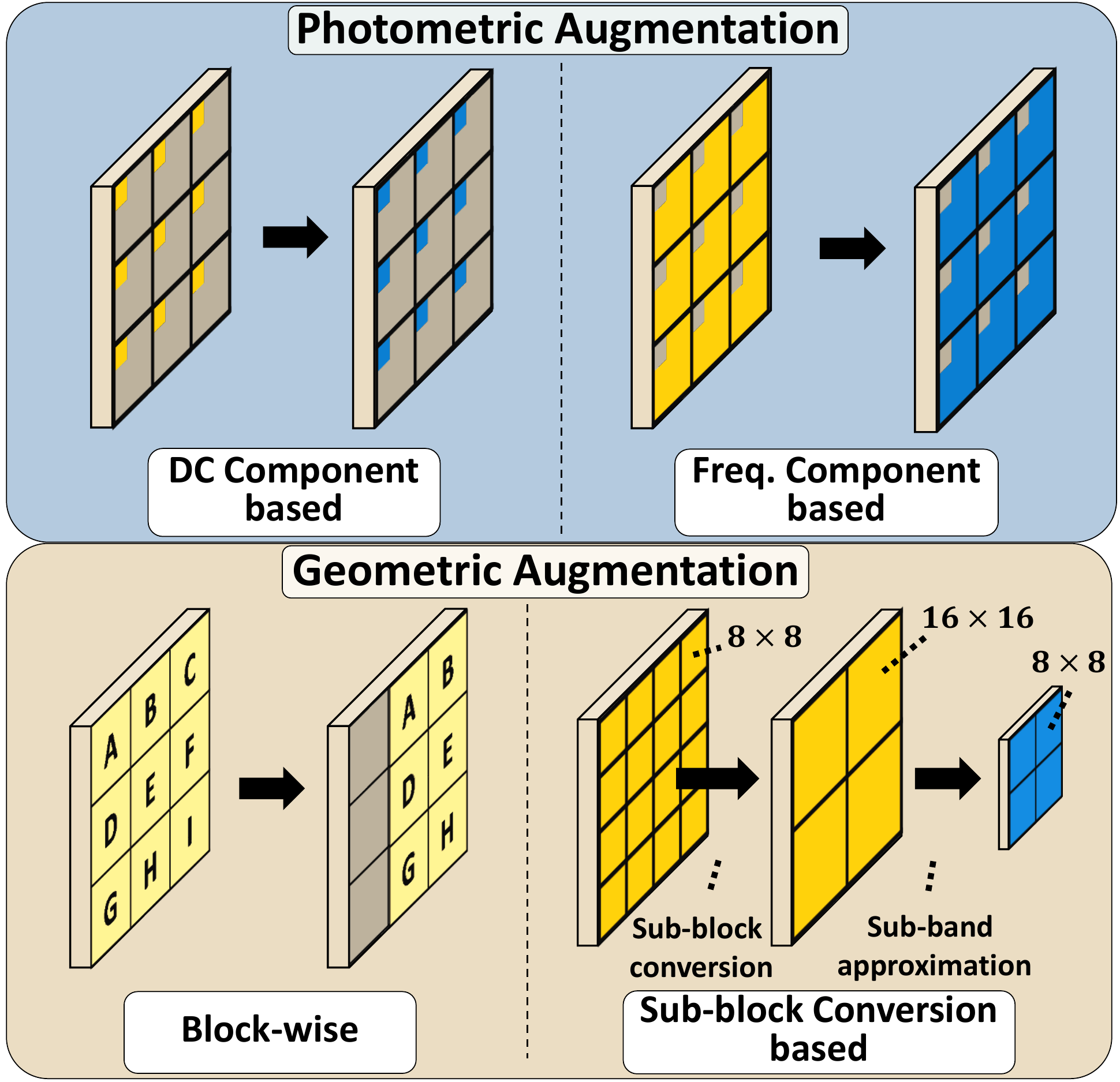}
    \caption{A visualization of different DCT augmentation types. Orange-colored coefficients are augmented and saved as blue. The bottom two examples illustrate \textit{Translate} and \textit{Resize}.}
    \label{fig:dctaug_process}
\end{figure}

\subsection{Photometric augmentation}
Photometric augmentation alters a metric of an image, which includes brightness or sharpness. Our implementation of this can be roughly categorized into two: DC component-based and frequency component-based augmentation. Both use the attributes of a DCT coefficient.
\label{aug:photometric}

\ourpar{DC Component-based augmentation} only alters the DCT coefficient without frequency ($X_{0,0}$), which is simply a scaled sum of all pixel values in the block (\cref{eq:dct8by8}). Altering this value will affect all pixels in the block evenly. This property can be used in two ways -- either when we have to modify a value uniformly across all pixels, or when we have to approximate a value of the pixels in a block.

\textit{Brightness} augmentation alters the brightness of the image. Considering that the image is represented in YCbCr color space, we can implement this augmentation by simply modifying the DC component of the Y channel. Let $Y_{u,v}^{h,w}$ denote a $u, v$-th DCT coefficient of Y at block position $h,w$. Then our implementation $f$ is the following where $t \in \mathbb{R}$.
\begin{equation}
    \resizebox{0.87\columnwidth}{!}{$
    f_{u, v}: Y_{u, v}^{h,w} \rightarrow \left\{\begin{array}{ll}
        Y_{0, 0}^{h,w} + t\cdot\texttt{mean}(\texttt{abs}(Y_{0,0})) & \text{if $u, v = 0$} \\
        Y_{u,v}^{h,w} & \text{otherwise}
    \end{array}\right\}$}
\end{equation}

\textit{Contrast} modifies the distance between the bright and dark values in the image. Since DCT coefficients are zero-centered by design (\cref{bak:jpegcomp} \labelcref{enum:zerocentered}) and the brightness of a block can be approximated using the DC component $Y_{0,0}$, we can categorize the non-negative component as `bright' and the negative component as `dark'. Therefore, multiplying $Y_{0,0}$ with $t \in [0, \infty)$ can adjust the contrast of the image. 
The implementation is $f: Y_{0,0}^{h,w} \rightarrow tY_{0,0}^{h,w}$. Similarly, \textit{Equalize} applies histogram equalization to $Y_{0,0}$.

\textit{Color (Saturation)} augmentation apply the same equation of \textit{Contrast} to $U$ (DCT coefficient of CbCr), instead of $Y$. This augments the saturation as illustrated in \cref{fig:randaugments}.

\textit{AutoContrast} scales the values so that the brightest and darkest value in an image becomes the brightest and darkest possible value. These values are approximated using $X_{0,0}$ in the same way as \textit{Contrast}. 
In \cref{bak:DCTscale}, we showed that the min/max value of the DCT in JPEG is $[-1024, 1016]$. Thus, our implementation $f$ is as follows.
\begin{equation}
    \resizebox{\columnwidth}{!}{$
    f: Y_{u, v}^{h,w} \rightarrow \left\{\begin{array}{ll}
        \frac{Y_{0, 0}^{h,w}-\min(Y_{0,0})}{\max(Y_{0,0})-\min(Y_{0,0})} \cdot 2040 - 1024 & \text{if $u, v = 0$} \\ 
        Y_{u, v}^{h,w} & \text{otherwise}
        \end{array}\right\}$}
\end{equation}

\textit{AutoSaturation$^*$} applies the formula in \textit{AutoContrast} to $U$. This allows the DCT to utilize the full range of the color.

\ourpar{Frequency component-based augmentation} uses non-DC components ($X_{u,v},\, u,v \neq 0$).
Altering the amplitude of these changes the intensity of the corresponding cosine signal in the pixel space. Here, we designed three augmentations each affecting frequency differently.

\textit{Sharpness} augmentation adjusts a sharpness of an image. Typically in RGB, this utilizes a convolution kernel to achieve such an effect. However, in DCT, several studies show that this can be implemented by adjusting the frequency components of the DCT \cite{dctblurdetection1,dctblurdetection2,dctblurdetection3,jpegsharpen1,jpegsharpen2,jpegsharpen3,jpegsharpen4}. They show that sharper images will generally have a higher frequency as there are more sudden changes around the sharp edges.
Using this property, we implement \textit{Sharpness} by linearly altering the frequency components. If $t>0$, the following equation sharpens the image. Otherwise, it blurs it.
\begin{equation}
    \resizebox{0.87\columnwidth}{!}{$
    f: Y_{u, v}^{h,w} \rightarrow Y_{u, v}^{h,w} \cdot \max(1+\frac{tu}{7}, 0)\max(1+\frac{tv}{7}, 0)
    $}
\end{equation}

\textit{MidfreqAug$^*$} augmentation is similar to \textit{sharpness} but instead of peaking the augmentation strength at the highest frequency $(u, v = 7)$, we peaked it at the middle frequency $(u, v\in\{3,4\})$. We expect the results to be similar to \textit{Sharpness}, but possibly with less noise.

\textit{FreqEnhance$^*$} multiplies all frequency components uniformly with a positive factor $t \in [0, \infty)$. The augmentation is simply $f: Y_{u,v}^{h,w} \rightarrow tY_{u,v}^{h,w},\; u,v \neq 0$. We believe that this allows us to see the impact of a frequency component with respect to the model performance.

\ourpar{Photometric -- special case.} There are some augmentations that do not categorize into either of the above augmentations. \textit{Invert} simply flips the sign of all DCT coefficients. This is because the coefficients are virtually zero-centered. \textit{Posterize} quantizes $X_{0,0}$ to lower bits. \textit{Solarize} uses $X_{0,0}$ to determine whether or not the DCT block should be inverted. \textit{SolarizeAdd} adds a preset value to $X_{0,0}$ if it is below threshold. \textit{Grayscale} replaces $U$ with zeros, removing color information. \textit{ChromaDrop$^*$} instead drops the Cb or Cr channel randomly, removing only half of the color information.

\subsection{Geometric augmentation}
Geometric augmentation modifies the image plane geometrically. An example of this augmentation includes translation, rotation, or shearing. There are two main sub-categories of geometric augmentation -- block-wise and sub-block conversion-based augmentation.

\ourpar{Block-wise augmentation} treats the DCT block positions similarly to pixel positions. \textit{Translate} can be implemented by moving the positions $h, w$ of each $X^{h,w}$ and filling the blank with zeros. \textit{Cutout} follows a similar process where we crop blocks out and fill them with zeros.

\textit{Flipping} the DCT coefficients utilize the fact that odd-column or odd-row DCT bases are odd-symmetric \cite{jpegfliprot90}. \textit{Flip} is performed by flipping the position of the blocks and then flipping the individual DCT blocks.
Define $R = \texttt{diag}(1, -1, 1, -1, ... -1)$ of matching size. Then, the per-block flipping operation is as follows.
\begin{equation}
    \label{eq:flip}
    f^X: X^{h,w} \rightarrow \bigg\{\begin{array}{ll}
        X^{h,w}R & \text{if horizontal flip} \\
        RX^{h,w} & \text{if vertical flip}
    \end{array}\bigg\}
\end{equation}

\textit{Rotate90} is implemented using a transpose with flipping \cite{jpegfliprot90}. We first rotate the block position $h, w$ and rotate each $X^{h,w}$ by 90 degrees. The per-block rotation is defined as:
\begin{equation}
    \label{eq:rot90}
    \resizebox{0.87\columnwidth}{!}{$
    \displaystyle f^X: X^{h,w} \rightarrow \bigg\{\begin{array}{ll}
        (X^{h,w})^T R & \text{clockwise} \\
        R(X^{h,w})^T & \text{counter-clockwise}
    \end{array}\bigg\}$}
\end{equation}

\label{subsec:subblock}
\ourpar{Sub-block conversion-based augmentation} uses the relationship between the DCT block and its smaller sub-blocks. This allows us to efficiently calculate the DCT of different DCT bases without the need to do inverse transform (e.g. $8 \times 8 $ DCT$ \leftrightarrow 32 \times 32$ DCT).
The relationship studied by Jiang and Feng \cite{subblock} is as follows. Let $X_{N\times M}^{h,w}$ be the DCT coefficient block of $N\times M$ DCT at block position $h, w$. \vspace{-2pt} 
Then, there exists a conversion matrix $A$ such that:
\begin{equation}
    \label{eq:subblockcomb}
    \resizebox{0.85\columnwidth}{!}{$
    X_{LN\times MN} = A_{L,N}\begin{bmatrix}
        X_{N\times N}^{0,0} & \cdots & X_{N\times N}^{0,M-1} \\
        \vdots & \ddots & \vdots \\
        X_{N\times N}^{L-1,0} & \cdots & X_{N\times N}^{L-1,M-1}
    \end{bmatrix}A^T_{M,N}$}
\end{equation}
\noindent Where $A_{L,N}$ is a $LN \times LN$ matrix that converts the $L$ number of $N$ 1-D DCT blocks into a single $LN$ DCT block. The decomposition of $X_{LN\times MN}$ DCT blocks into $L\times M$ DCT blocks of $X_{N\times N}$ follows a similar process:
\begin{equation}
    \label{eq:subblockdecomp}
    \resizebox{0.85\columnwidth}{!}{$
    \begin{bmatrix}
        X_{N\times N}^{0,0} & \cdots & X_{N\times N}^{0,M-1} \\
        \vdots & \ddots & \vdots \\
        X_{N\times N}^{L-1,0} & \cdots & X_{N\times N}^{L-1,M-1}
    \end{bmatrix} = A_{L,N}^{-1} X_{LN\times MN} A^{-1^T}_{M,N}$}
\end{equation}
Derivation of $A$ is given in the \cref{apdx:convmat}.

\textit{Resize} can be implemented if we can understand how to resize individual DCT blocks. Suppose that there exists a way to resize $X_{4 \times 4}$ to $X_{8 \times 8}$ by padding. Then, to upsample $X_{8 \times 8}$ while keeping the $8\times 8$ sliced structure of JPEG DCT, we can first decompose $X_{8 \times 8}$ into four $X_{4 \times 4}$ using \textit{sub-block conversion}. Then, we can individually resize each $X_{4 \times 4}$ to $X_{8 \times 8}$. This gives us four $X_{8 \times 8}$ blocks upsampled from one $X_{8\times 8}$. Downsampling can follow a similar process. We first combine four adjacent $X_{8 \times 8}$ into a single $X_{16 \times 16}$ using \textit{sub-block conversion}. Then, we resize $X_{16 \times 16}$ down to $X_{8\times 8}$. This process is shown in \cref{fig:dctaug_process}.

This technique to resize each individual DCT block is known as \textit{sub-band approximation} and has been studied by Mukherjee and Mitra \cite{dctresizing}. Their work shows that
if $X_{N\times N}(k,l)$ is a $(k,l)$-th coefficient of a $X_{N \times N}$ block, then, the approximate relationship is:
\begingroup
\begin{equation}
    \label{eq:subbandapprox_lm}
    \resizebox{\columnwidth}{!}{$
    X_{LN \times MN}(k,l) \approx \bigg\{ \begin{array}{ll}
         \sqrt{LM} \; X_{N \times N}(k,l) & 0 \leq k,l \leq N-1  \\
         0 & \text{otherwise}
     \end{array}$}
\end{equation}
\endgroup
Using this, we can upsample $L \times M$ times by resizing each $X_{N \times N}$ to $X_{LN \times MN}$ and decomposing them to $L \times M$ $X_{N \times N}$. $L \times M$ downsampling combines $L \times M$ adjacent $X_{N \times N}$ to form $X_{LN \times MN}$ and resize it to $X_{N \times N}$. An arbitrary resizing of $\frac{P}{Q} \times \frac{R}{S}$ can be done by first upsampling $P \times R$ times and downsampling the result by $Q \times S$.

\algdef{SE}[SUBALG]{Indent}{EndIndent}{}{\algorithmicend\ }%
\algtext*{Indent}
\algtext*{EndIndent}

\begin{table*}[ht!]
    \renewcommand\cellset{\renewcommand\arraystretch{0.8}%
    \setlength\extrarowheight{0pt}}
    \centering
    \renewcommand{\arraystretch}{1.0}
    \resizebox{0.95\textwidth}{!}{
    \begin{tabular}{c c c|c|cc|c c c | c c c|c}
        Architecture & \makecell[b]{Color\\Space} & \makecell[b]{Aug.\\Space} & \makecell[b]{Embed.\\FLOPs} & Decode & Augment & \makecell[b]{Train\\Data Load} & \makecell[b]{Model\\Fwd/Bwd} & \makecell[b]{Train\\Pipeline} & \makecell[b]{Eval\\Data Load} & \makecell[b]{Model\\Fwd} & \makecell[b]{Eval\\Pipeline} & \makecell[b]{Val\\Acc (\%)}\\ 
        \hline
        \multicolumn{3}{c|}{\footnotesize ($\downarrow$ Performance)} & & \multicolumn{2}{c|}{\footnotesize Latency per img (ms)} & \multicolumn{6}{c|}{\footnotesize Throughput per GPU (FPS)} &\\
        ViT-Ti \cite{betterbaselinevit} & RGB & RGB & 28.9M & 3.68 & 3.08 &  571.5 &  832.8 &  493.8 & 641.4 & 2898.7 & 638.0 & 74.1 \\ 
        ViT-S \cite{betterbaselinevit} & RGB & RGB & 57.8M & 3.62 & 3.01 &  558.0 &  355.7 &  352.1 & 660.2 & 1174.5 & 610.5 & 76.5 \\
        ViT-S$^{\bigstar}$ \cite{betterbaselinevit} & RGB & RGB & 57.8M & 3.69 & 2.63 &  574.7 &  716.8 &  489.0 & 680.3 & 2335.1 & 644.2 & 75.6 \\
        SwinV2-T$^{\bigstar}$ \cite{liu2021swinv2} & RGB & RGB & 20.8M & 3.61 & 2.98 & 489.8 & 231.5 & 231.6 & 614.8 & 809.5 & 516.2 & 79.0$^\star$\\
        ResNet50\cite{torchvision} & RGB & RGB & - & - & - & - & - & - & 688.2 & 1226.8 & 639.1 & 76.1 \\ 
        \hline
        JPEG-Ti & DCT & DCT & 16.1M & 1.42 & 2.62 &  816.2 &  857.2 &  687.6 & 775.3 & 2847.5 & 752.3 & 75.1 \\ 
        JPEG-S & DCT & DCT & 30.5M & 1.49 & 2.40 &  824.8 &  364.3 &  360.5 & 782.9 & 1139.7 & 711.1 & 76.5 \\
        JPEG-S$^{\bigstar}$ & DCT & DCT & 30.5M & 1.42 & 2.37 &  821.7 &  764.0 &  665.8 & 793.1 & 2384.1 & 711.8 & 75.8 \\
        JPEG-S$^{\bigstar\diamondsuit}$ & DCT & \makecell[c]{\footnotesize RGB to\\\footnotesize DCT} & 30.5M & 3.62 & 5.98 & 425.3 & 751.9 & 372.3 & 748.7 & 2382.4 & 721.0 & 76.3 \\
        SwinV2-T$^{\bigstar}$ & DCT & DCT & 13.0M & 1.40 & 2.47 & 752.0 & 241.9 & 235.9 & 664.2 & 824.9 & 578.0 & 79.4\\
        ResNet50$^{\diamondsuit}$\cite{fasternnjpeg} \cite{learninginfreq} & DCT & \makecell[c]{\footnotesize RGB to\\\footnotesize DCT} & - & - & - & - & - & - & 785.7 & 5969.8 & 753.5 & 76.1\\ 
        \hline
        \multicolumn{3}{c|}{\footnotesize ($\downarrow$ Improvements)} & \multicolumn{3}{c|}{\footnotesize (Reduction $\blacktriangledown$)} & \multicolumn{6}{c|}{\footnotesize (Speed-ups $\blacktriangle$)} &\\
        \multicolumn{3}{c|}{JPEG-Ti vs ViT-Ti} & -44.4 \% & -61.4 \% & -14.9 \% &  +42.8 \% &  +2.9 \% &  \textbf{+39.2} \% & +20.9 \% & -1.8 \% & \textbf{+17.9 \%} & +1.0\\
        \multicolumn{3}{c|}{JPEG-S vs ViT-S} & -47.2 \% & -58.8 \% & -20.3 \% &  +47.8 \% &  +2.4 \% &  +2.4 \% & +18.6 \% & -3.0 \% & +16.5 \% & +0.0\\
        \multicolumn{3}{c|}{JPEG-S$^{\bigstar}$ ViT-S$^{\bigstar}$} & -47.2 \% & -61.5 \% & -9.9 \% &  +43.0 \% &  +6.6 \% &  +36.2 \% & +16.6 \% & +2.1 \% & +10.5 \% & +0.2\\
        \multicolumn{3}{c|}{JPEG-S$^{\bigstar}$ vs JPEG-S$^{\bigstar\diamondsuit}$} & +0.0 \% & -60.8 \% & -60.4 \% &  +93.2 \% &  +1.6 \% &  +78.8 \% & +5.9 \% & +0.1 \% & -1.3 \% & -0.5\\
        \multicolumn{3}{c|}{SwinV2-T$^{\bigstar}$ (DCT vs RGB) } & -37.7 \% & -61.2 \% & -17.1 \% &  +53.5 \% &  +4.5 \% &  +1.9 \% & +8.0 \% & +1.9 \% & +12.0 \% & +0.4\\
    \end{tabular}
    }
    \caption{Model throughput per GPU for each pipeline element and accuracy.
    \textit{Embed. FLOPs} shows the FLOPs needed to generate patch embeddings. \textit{Decode} and \textit{Augment} indicates the per-image processing latency during training. 
    \textit{Val Acc} shows the accuracy on the ImageNet validation set. `JPEG-' prefix indicates that it is a ViT trained using JPEG DCT coefficients. 
    Model with `$\bigstar$' symbol is trained using mixed precision~\cite{micikevicius2018mixed}. `$\diamondsuit$' models are trained using the pipeline suggested by Gueguen \etal \cite{fasternnjpeg}. $^\star$SwinV2 models are trained using the same recipe as in \cite{betterbaselinevit} for a fair comparison. The details of these measurements are shown in \cref{apdx:measurement}.}
    \label{tab:benchmark_and_acc}
\end{table*}

\textit{Rotate} is implemented using the rotational property of the Fourier transform \cite{bracewellfourieraffine,bernardini2000fourierrotation,o1997fourierrotation2}. This property denotes that the Fourier transform of a rotated function is equal to rotating the Fourier transform of a function. 
To use this property, we slightly alter the \cref{eq:subblockcomb}. Instead of combining the blocks to $X_{LN\times MN}$ DCT, we combine them to the discrete Fourier transform (DFT) coefficients. Define $D_{N \times N}$ as the DFT coefficient block of size $N \times N$. Then, the rotation is done by combining $L \times M$ $X_{N \times N}$ to $D_{LN \times MN}$, rotating it, and decomposing it back to $L \times M$ $X_{N \times N}$ using the modified \cref{eq:subblockdecomp}. This can be further improved using the lossless 90-degree rotation to minimize the lossy arbitrary-degree rotation. The details of this DFT conversion are shown in \cref{apdx:fourierdecomp}.

\textit{Shear} is identically implemented as \textit{Rotate} where instead of rotating $D_{LN \times MN}$, we shear it. This is because we can apply the same property \cite{bracewellfourieraffine} for each axis.

\section{Experiments}
\label{sec:experiments}

In this section, we compare our models trained in DCT space with the RGB models. We show that the DCT models achieve similar accuracy but perform notably faster than the RGB models.
First, we compare the throughput and accuracy of the RGB and DCT models. Then, we compare the DCT embedding strategies covered in \cref{sec:ModelArch}. Lastly, we conduct an ablation study on the DCT data augmentation.

\newcommand{\cmark}{\color{BlueGreen}\ding{51}} 
\newcommand{\xmark}{{\color{WildStrawberry}\ding{55}}} 
\begin{table*}[th!]
    \centering
    \resizebox{0.95\textwidth}{!}{
    \begin{tabular}{c|c|ccc|cc|c|c}
         & & \multicolumn{3}{c|}{Photometric} & \multicolumn{2}{c|}{Geometric} & \multicolumn{2}{c}{Accuracy} \\
        \hline
        Subset & Resize & DC based & Freq. based & Special case & Block-wise & Sub-block based & \makecell[c]{RGB} & \makecell[c]{DCT} \\
        \hline
        \footnotesize (Avg. latency (ms) $\downarrow$) & & & & & & & &\\
        RGB & 2.17 & 0.61 & 1.34 & 0.67 & 1.23 & 1.28 & - & -\\
        DCT & 2.79 & 0.37 & 0.54 & 0.34 & 0.42 & 9.31 & - & - \\
        \hline
        \footnotesize(Ablation $\downarrow$) & & & & & & & &\\
        All & \cmark & \cmark & \cmark & \cmark & \cmark & \cmark & 76.5 & 74.6\\
        No sub-block & \cmark & \cmark & \cmark & \cmark & \cmark & \xmark & 76.2 & 75.1\\
        No block-wise & \cmark & \cmark & \cmark & \cmark & \xmark & \xmark & 76.0 & 72.6\\
        No Special-case & \cmark & \cmark & \cmark & \xmark & \xmark & \xmark & 76.3 & 73.3\\
        No Freq. based &\cmark & \cmark & \xmark & \xmark & \xmark & \xmark & 75.9 & 74.2\\
        No DC based & \cmark & \xmark & \xmark & \xmark & \xmark & \xmark & 74.4 & 75.5\\
        \hline
        Best Subset & \cmark & \multicolumn{4}{c|}{\makecell[c]{\fontsize{8}{10}\selectfont Brightness, Contrast, Color, AutoContrast, AutoSaturation, Sharpness, \\ \fontsize{8}{10}\selectfont MidfreqAug, Posterize, Grayscale, ChromaDrop, Translate, Cutout, Rotate90}} & \xmark & 76.5$^{\star}$ & 76.5
    \end{tabular}
    }
    \caption{Data augmentation ablation study on ViT-S. 
    At the bottom row, we report the best subset we found and train both the RGB and DCT models from it. $^\star$RGB accuracy is obtained with \textit{MidfreqAug} removed, since it is analogous to \textit{Sharpness} in RGB as shown in \cref{fig:randaugments}.
    }
    \label{tab:aug_ablation}
\end{table*}

\begin{table}[th!]
    \centering
    \resizebox{0.8\columnwidth}{!}{
    \begin{tabular}{ccc|c|c}
        Grouped & Separate & Concat. & Sub-block & \makecell[c]{Acc} \\
        \hline
        \cmark & -  & - & \cmark & \textbf{76.5}\\
        - & \cmark & - & \cmark & 74.7\\
        - & - & \cmark & \cmark & 71.3\\
        \cmark & - & - & - & 74.6\\
        - & \cmark & - & - & 73.8\\
        - & - & \cmark & - & 71.3\\
    \end{tabular}
    }
    \caption{Embedding strategy ablation study on ViT-S. We see that grouped embedding with sub-block outperforms other strategies.}
    \label{tab:modelablation}
\end{table}

\ourpar{Implementation Details.} All experiments are conducted with PyTorch~\cite{paszke2019pytorch}.
We extract DCT coefficients using a modified version of \texttt{libjpeg}~\cite{libjpeg} and TorchJPEG~\cite{torchjpeg}.
All timing measurements are performed using $2\times$ A40 GPUs and 8 cores from an Intel Xeon 6226R CPU, and all throughputs are reported per GPU.
We used \textit{fvcore}~\cite{fvcore} to obtain the FLOPs.
All models are trained on ImageNet~\cite{imagenet,russakovsky2015imagenet}, which we resize on-disk to $512\times512$ prior to training.
We re-implement a ViT training pipeline in PyTorch, carefully following the recipe suggested by Beyer \etal \cite{betterbaselinevit} which uses random resized crop, random flip, RandAugment~\cite{randaugment} and Mixup~\cite{mixup} with a global batch size of 1024. All plain ViT models are trained with a patch size of 16.
SwinV2 model \cite{liu2021swinv2} uses a patch size of 4, a window size of 8, and a global batch size of 512.
`-S' models are trained for 90 epochs. `-Ti' and SwinV2 models are instead trained for 300 epochs using the same recipe in \cite{betterbaselinevit}. 
Following \cite{betterbaselinevit}, we randomly sample 1\% of the train set to use for validation.
Our RGB ResNet-50 baseline uses the V1 weights from Torchvision~\cite{torchvision}.
There exist recent works with improved training recipes for ResNets~\cite{wightman2021resnet,bello2021revisiting},
but this is orthogonal to our work. The DCT ResNet-50 uses the model proposed by Gueguen \etal~\cite{fasternnjpeg} and Xu \etal~\cite{learninginfreq}.

\subsection{Main results}
The hyperparameter settings for the models are given in \cref{apdx:trainingsettings}.
\cref{tab:benchmark_and_acc} shows the comprehensive result from our experiment. 
Both the ViT and CNN models show equivalent accuracy to their RGB counterparts. However, loading directly from JPEG DCT reduces decoding latency by up to 61.5\%. We see a similar effect on the augmentation latency to a lesser extent. Additionally, as the input size has been halved by JPEG (\cref{bak:jpegcomp}), we observe that the FLOPs needed to generate embeddings are reduced by up to 47.2\%.
These speed-ups boost throughput for DCT models. Most notably, our JPEG-Ti model showed 39.2\% and 17.9\% faster training and evaluation.
Our augmentation scheme improves train data loading throughput by 93.2\% versus the one by Gueguen \etal \cite{fasternnjpeg}, which must convert DCT to RGB, augment, and convert it back to DCT.

While training as-is still reaps the benefits of faster data loading, we can observe that the JPEG-S model is bottlenecked by model forward and backward passes. One option is to employ mixed precision training \cite{micikevicius2018mixed}. This allows us to fully realize the data loading speed-ups by accelerating the model with minor accuracy loss. We observe that our JPEG-S model trained using mixed precision is 36.2\% faster during training time compared to the RGB counterpart. We believe most other faster models discussed in \cref{sec:related} will also benefit from this speed-up.

\subsection{Embedding strategies}
\label{subsec:modelarch}
To see which patch embedding scheme is most suitable for DCT, we performed an ablation study in \cref{tab:modelablation}. We ablate on the embedding strategies and the sub-block conversion technique discussed in \cref{sec:ModelArch} on ViT-S. The results show two things. One is that the simplest strategy--grouped embedding--is best, and that sub-block conversion is important to achieve high accuracy. We believe this indicates that (1) all blocks in an image patch relay information of that patch as a whole; they should not be separately inferred, and (2) it is more natural for the model to deduce the information in a patch if the sub-blocks are fused together. The best strategy is used throughout the DCT models in \cref{tab:benchmark_and_acc}.

\subsection{Augmentation}
\label{subsec:augmentation}
As discussed in \cref{sec:DCTAug}, data augmentation is crucial to train a well-performing model. However, many existing studies \cite{dataaug1,dataaug2,dataaug3,dataaug4,dataaug5,dataaug6-autoaug,vit_datahungry,randaugment} are tuned toward RGB augmentations; we cannot assume that the existing work's result would be consistent on DCT. In other words, some DCT augmentations could have a different impact on both the accuracy and throughput of the model compared to RGB. Therefore, we perform an ablation study on data augmentations discussed in \cref{sec:DCTAug} and report the result in \cref{tab:aug_ablation}. 
We perform ablation by varying the subset of augmentations used in RandAugment \cite{randaugment} on ViT-S. The experiment is performed separately for RGB and DCT with their respective augmentation sets. The results show that while sub-block-based augmentations (e.g. \textit{rotate, shear}) are prohibitively expensive to do directly on DCT, it is not as important to train a model. We report the best augmentation subset we found without sub-block-based augmentations at the bottom row. In addition, we train an RGB model using this subset and compare the accuracy. We can see that the RGB model performs identically to the DCT model, and the subset does not unfairly favor DCT over RGB models.

\section{Conclusion}
\label{sec:conclusion}
In this paper, we demonstrated that vision transformers can be accelerated significantly by directly training from the DCT coefficients of JPEG. We proposed several DCT embedding strategies for ViT, 
as well as reasonable augmentations that can be directly carried out on the DCT coefficients. The throughput of our model is considerably faster with virtually no accuracy loss compared to RGB.

RGB has been widely used for as long as computer vision existed. Many computer vision schemes regard retrieving RGB as a necessary cost; they have not considered the potential of direct encoded-space training in-depth. We wanted to show the capability of this approach, as well as demonstrate that recovering RGB is not always necessary.

Encoded-space training can be adapted to nearly all scenarios that require loading some compressed data from storage. From mobile devices to powerful data centers, there are no situations where they wouldn't benefit from faster loading speed. As we only considered the plain ViT architecture, there is still more to be gained by adapting our findings to the existing efficient architectures. Future studies may also consider a better augmentation strategy to improve the performance further.
We hope that the techniques shown in this paper prove to be an adequate foundation for future encoded-space research.


{\small
\bibliographystyle{unsrt}
\bibliography{egbib}
}

\newpage
    \begin{appendices}
\crefalias{section}{appendix}
\raggedbottom
\captionsetup[table]{skip=3pt}
\label{sec:appendix}
\section{Scaling property of a $8\times 8$ DCT}
\label{apdx:scalingDCT}
Consider an $8 \times 8$ 2D-DCT defined as the following where $\alpha_i = 1/\sqrt{2}$ if $i=0$, else $1$, $u,v,m,n \in [0..7]$.
\begin{equation}
\hskip-10pt\resizebox{0.87\columnwidth}{!}{$
\displaystyle X_{u,v} = \frac{\alpha_u \alpha_v}{4} \sum_{m,n} x_{m,n} \cos\big[\frac{\pi(2m+1)u}{16}\big]\cos\big[\frac{\pi(2n+1)v}{16}\big]
$}
\label{eq:dct8by8_apdx}
\end{equation}
We then calculate how much the DCT scales up the min/max values considering the following two cases.
\begin{itemize}[leftmargin=4pt,itemindent=5pt]
    \item If $u, v = 0$\vspace{0.1cm}\\
    If $u, v = 0$ then the resulting coefficient $X_{0,0}$ is simply:
    \begin{align}
        X_{0,0} &= \frac{1}{8}\sum_{m=0}^{7} \sum_{n=0}^{7} x_{m,n} \cdot \cos(0) \cos(0)\\
        &= \frac{1}{8}\sum_{m=0}^{7} \sum_{n=0}^{7} x_{m,n} \hskip 0.3cm (\because \cos(0)=1)
    \end{align}
    Which is minimized/maximized when $x_{m,n}$ is min/max:
    \begin{align}
        \min(X_{0,0}) &= \frac{1}{8} \min(x_{m,n}) \cdot 8 \cdot 8\\
        &= \min(x_{m,n}) \cdot 8\\
        \max(X_{0,0}) &= \frac{1}{8} \max(x_{m,n}) \cdot 8 \cdot 8\\
        &= \max(x_{m,n}) \cdot 8
    \end{align}
    So applying $8\times8$ DCT to an input with min/max value of [-128, 127] scales the output min/max value to:
    \begin{align}
        \min(X_{0,0}) &= -128 \cdot 8 = -1024\\
        \max(X_{0,0}) &= 127 \cdot 8 = 1016         
    \end{align}
    
    \item If $u \neq 0$ or $v \neq 0$ \vspace{0.1cm}\\
    In this case, the amplitude of DCT coefficient $X_{u, v}$ will be maximized if the input data sequence resonates with (i.e. match the signs of) the cosine signal of the corresponding DCT bases. We will show that this value is less than or equal to the magnitude when $u, v = 0$. We first consider the 1D case and then use it to calculate the 2D case. The 1D DCT is as follows.
    \begin{align}
        X_{u} &= \frac{\alpha_u}{2} \sum_{m=0}^{7} x_m \times \cos\bigg[\frac{\pi(2m+1)u}{16}\bigg]
    \end{align}
    Where the DCT bases are:
    \begin{align}
        \frac{\alpha_u}{2}\cos\bigg[\frac{\pi(2m+1)u}{16}\bigg], \hskip0.3cm m, u \in [0..7]
    \end{align}
    This 1D DCT will be maximized when the signs of $x_m$ match the signs of the DCT bases. Likewise, it will be minimized when the signs are exactly the opposite. Therefore, we compare the absolute sum of the DCT bases and show that it is less than or equal to the sum when $u=0$. This absolute sum of DCT bases can be interpreted as the \textit{scale-up factor} as it shows how much the input 1's with matching signs are scaled up. The following values are rounded to the third decimal place.
    \begin{itemize}
        \setlength{\itemindent}{-0.2cm}
        \item $u=0$: {\footnotesize $\displaystyle X_0 = \frac{\alpha_0}{2}\sum_{m=0}^{7} \texttt{abs}(1) = 2\sqrt{2} = \mathbf{2.828}$}
        \item $u=1$: {\footnotesize $\displaystyle X_1 = \frac{\alpha_1}{2}\sum_{m=0}^{7} \texttt{abs}(\cos\bigg[\frac{\pi(2m+1)}{16}\bigg]) = 2.563$}
        \item $u=2$: {\footnotesize $\displaystyle X_2 = \frac{\alpha_2}{2}\sum_{m=0}^{7} \texttt{abs}(\cos\bigg[\frac{2\pi(2m+1)}{16}\bigg]) = 2.613$}
        \item $u=3$: {\footnotesize $\displaystyle X_3 = \frac{\alpha_3}{2}\sum_{m=0}^{7} \texttt{abs}(\cos\bigg[\frac{3\pi(2m+1)}{16}\bigg]) = 2.563$}
        \item $u=4$: {\footnotesize $\displaystyle X_3 = \frac{\alpha_4}{2}\sum_{m=0}^{7} \texttt{abs}(\cos\bigg[\frac{4\pi(2m+1)}{16}\bigg]) = 2.828$}
        \item $u=5$: {\footnotesize $\displaystyle X_3 = \frac{\alpha_5}{2}\sum_{m=0}^{7} \texttt{abs}(\cos\bigg[\frac{5\pi(2m+1)}{16}\bigg]) = 2.563$}
        \item $u=6$: {\footnotesize $\displaystyle X_3 = \frac{\alpha_6}{2}\sum_{m=0}^{7} \texttt{abs}(\cos\bigg[\frac{6\pi(2m+1)}{16}\bigg]) = 2.613$}
        \item $u=7$: {\footnotesize $\displaystyle X_3 = \frac{\alpha_7}{2}\sum_{m=0}^{7} \texttt{abs}(\cos\bigg[\frac{7\pi(2m+1)}{16}\bigg]) = 2.563$}
    \end{itemize}
    We can see that for all $u$, the absolute sums of DCT bases are less than or equal to the sum when $u=0$. 2D DCT is simply a DCT on each axis (rows and columns), so the 2D scale-up factors will be a pairwise product for any pairs of $u$ with replacement. This will still not exceed the value when we choose $u=0$ twice. Therefore, we can conclude that the minimum and maximum values calculated in the $u,v=0$ case will hold for all $8\times 8$ DCT coefficients. Thus, $8 \times 8$ DCT will scale up the min/max value by 8.
\end{itemize}

\section{Compute cost to decode JPEG}
\label{apdx:jpegcost_detail}
JPEG is decoded through the following steps.
\begin{enumerate}[label=(\alph*)]
    \itemsep-3pt
    \item Decode Huffman codes to RLE symbols
    \item Decode RLE symbols to quantized DCT coefficients
    \item De-quantize the DCT coefficients
    \item Apply inverse-DCT to the DCT coefficients
    \item Shift value from [-128, 127] to [0, 255]
    \item Upsample Cb, Cr by $2\times$ for each dimension
    \item Convert YCbCr to RGB
\end{enumerate}
We count the number of operations (OPs) for each step:
\begin{enumerate}[label=(\alph*)]
    \itemsep-3pt
    \item Read $N_s$ Huffman codes and recover $N_s$ RLE symbols = $N_s + N_s = 2N_s$ OPs
    \item Read $N_s$ RLE symbols and recover $8 \times 8$ quantized DCT coefficients = $N_s + 64$ OPs
    \item Multiply $8 \times 8$ quantization table element-wise with $8 \times 8$ DCT coefficients = $64$ OPs
    \item The $8 \times 8$ inverse DCT is given as:
    \begin{equation}
        \hskip-10pt\resizebox{0.82\columnwidth}{!}{$
        \displaystyle x_{m, n} = \frac{1}{4}\gamma_{m}\gamma_{n}\sum_{u,v}X_{u, v}\cos\bigg[\frac{(2m+1)u\pi}{16}\bigg]\cos\bigg[\frac{(2n+1)v\pi}{16}\bigg]
        $}
    \end{equation}
    Where
    \begin{align*}
        \gamma_{i} &= \Bigg\{\begin{array}{ll}
            \frac{1}{\sqrt{2}} & \text{if }i=0 \vspace{0.1cm}\\
            1 & \text{otherwise}
        \end{array}
    \end{align*}
    \begin{itemize}[leftmargin=10pt]
        \itemsep -3pt
        \item If $m, n \in [1..7]$\\
        We need $2 \cos + 10 \text{mul} + 3 \text{add} + 3 \text{div}$ per step as $\gamma_{i} = 1$. Number of OPs: $7\times 7 \times 18 = 882$
        \item If $m = 0, n \in [1..7]$ or $m \in [1..7], n = 0 $\\
        We need $2 \cos + 10 \text{mul} + 3 \text{add} + 4 \text{div} + 1 \text{sqrt}$ per step. OPs: $7 \times 20 \times 2 = 280$
        \item If $m =0, n = 0$\\
        $2 \cos + 10 \text{mul} + 3 \text{add} + 5 \text{div} + 2 \text{sqrt}$ = $22$ OPs
    \end{itemize}\vspace{-9pt}
    Total OPs per $8 \times 8$ block: $882 + 280 + 22 = 1184$
    \item Add 128 to every elements of $x_{m, n}$ = 64 OPs
    \item Upsample $8\times 8$ Cb and Cr block to $16 \times 16$: $256 \times 2=512$ OPs per 6 blocks. This is because 4 Y blocks are paired with 1 Cb and Cr block.
    Per-block cost: $512 / 6 = 85.3$ OPs.
    \item YCbCr is converted to RGB using \cite{jpegjfifrec}: 
    \begin{align*}
        R &= Y + 1.402(C_r - 128)\\
        G &= Y - 0.344136(C_b - 128) - 0.714136(C_r - 128)\\
        B &= Y + 1.772(C_b - 128)
    \end{align*}
    For three blocks -- Y, Cb, and Cr -- the number of OPs is $64 \times (2 \text{add} + 6 \text{sub} + 4 \text{mul}) = 768$. We ignore the cost of rounding and min/max clamping for simplicity. Thus, the per-block cost is $768/3 = 256$ OPs
\end{enumerate}
Recovering DCT coefficients requires going through steps (a)-(c), where the compute cost sums up to $3N_s + 128$ OPs. Full decoding requires $3N_s + 1717.3$ OPs. We can see that most of the decoding cost comes from the inverse-DCT, which costs 1184 OPs to compute.
Note that this result is only an estimate and can vary under different settings.

\section{Conversion matrix for sub-block conversion}
\label{apdx:convmat}
The conversion matrix $A$ can be calculated using the basis transform from $L\times M$ number of $N\times N$ DCT bases to $LN \times MN$ DCT bases. Let $T^{N\times N}$ as a 1-D DCT bases of size $N \times N$ then:
\begin{multline}
    \hskip-9pt\resizebox{0.88\columnwidth}{!}{$
    \displaystyle T^{N\times N} = \sqrt{\frac{2}{N}}\begin{bmatrix}
        \frac{1}{\sqrt{2}} & \frac{1}{\sqrt{2}} & \cdots & \frac{1}{\sqrt{2}}\\
        \cos[\frac{1\pi}{2N}] & \cos[\frac{3\pi}{2N}] & \cdots & \cos[\frac{(2N-1)\pi}{2N}]\\
        \vdots & \vdots & \ddots & \vdots \\
        \cos[\frac{(N-1)\pi}{2N}] & \cos[\frac{3(N-1)\pi}{2N}] & \cdots & \cos[\frac{(2N-1)(N-1)\pi}{2N}]
    \end{bmatrix}$}\hskip-3pt
\end{multline}

\noindent $T^{N\times N}$ is an orthogonal matrix \cite{dctorthogonal}. Hence, 
\begin{equation}
    \label{eq:dct_trans_inv}
    TT^T = I \hskip0.5cm T^T = T^{-1}
\end{equation}

\noindent Define $B_{large}$ as $T^{LN \times LN}$ and $B_{small}$ as a block diagonal matrix of $T^{N \times N}$ with size $LN \times LN$:

\begin{equation}
    \label{eq:bsmall_mat_def}
    B_{small} = \begin{bmatrix}
        T^{N \times N} & \cdots & 0\\
        \vdots & \ddots & \vdots\\
        0 & \cdots & T^{N \times N}
    \end{bmatrix}
\end{equation}

\noindent Then the conversion matrix $A_{L, N}$ is\cite{subblock}:
\begin{align}
    &B_{large} = A_{L, N} \times B_{small}\\
    &A_{L, N} = B_{large} \times B^{-1}_{small}
\end{align}

\noindent Where $B_{small}^{-1} = B_{small}^T$ due to \cref{eq:dct_trans_inv,eq:bsmall_mat_def}. Thus,
\begin{equation}
    \label{eq:conversion_matrix}
    A_{L,N} = B_{large} \times B_{small}^T
\end{equation}

\noindent We can also see that $B_{large}^{-1} = B_{large}^{T}$. Thus,
\begin{align}
    A_{L,N}^{-1} &= (B_{large} \times B_{small}^T)^{-1}\\
    &= (B_{small}^{T})^{-1} \times B_{large}^{-1}\\
    &= B_{small} \times B_{large}^T\\
    &= A_{L,N}^T
\end{align}

\section{Sub-band approximation}
\label{apdx:subbandapprox}
Define $x(m,n)$ as the 2D image data, and $X(k,l)$ as the 2D DCT coefficient of $x(m,n)$ where $m,n,k,l\in[0..N-1]$. Then, define $x_{LL}(m',n')$ as the $2 \times$ downsized image of $x(m,n)$. Then $x_{LL}$ is given as: 
\begin{multline}
    x_{LL}(m',n') = \frac{1}{4}\{x(2m',2n')+x(2m'+1,2n')+\\x(2m',2n'+1)+x(2m'+1,2n'+1)\}
\end{multline}
where $m',n',k',l' \in [0,..\frac{N}{2}-1]$.
Similarly, define $\overline{X_{LL}}(k',l')$ as the 2D DCT coefficient of $x_{LL}(m',n')$.
Mukherjee and Mitra's work \cite{dctresizing} shows that $X(k,l)$ can be represented in terms of $\overline{X_{LL}}(k,l)$:
 \begin{equation}
    \resizebox{0.88\columnwidth}{!}{$
     X(k,l) = \bigg\{ \begin{array}{ll}
         2\cos(\frac{\pi k}{2N})\cos(\frac{\pi l}{2N}) \overline{X_{LL}}(k,l) & 0 \leq k,l \leq \frac{N}{2}-1  \\
         0 & \text{otherwise}
     \end{array}$}
 \end{equation}

\noindent Which can be further simplified assuming that $k, l$ are negligible compared to $2N$: $\frac{\pi k}{2N}, \frac{\pi l}{2N} \approx 0$
\begingroup
\def\arraystretch{1.2}
\begin{equation}
\label{eq:subbandapprox_2x}
    X(k,l) \approx \bigg\{ \begin{array}{ll}
         2 \overline{X_{LL}}(k,l) & 0 \leq k,l \leq \frac{N}{2}-1  \\
         0 & \text{otherwise}
     \end{array}
\end{equation}
\endgroup

\noindent We can follow the same process for $L \times M$ downsampling from $LN \times MN$ DCT coefficient to $N \times N$ DCT \cite{dctresizing}:
\begingroup
\def\arraystretch{1.2}
\begin{equation}
\label{eqapdx:subbandapprox_lm}
\hskip-10pt\resizebox{0.87\columnwidth}{!}{$
    X(k,l) \approx \bigg\{ \begin{array}{ll}
         \sqrt{LM} \; \overline{X_{LL}}(k,l) & 0 \leq k,l \leq N-1  \\
         0 & \text{otherwise}
     \end{array}
     $}
\end{equation}
\endgroup

\noindent Thus, \cref{eqapdx:subbandapprox_lm} implies the approximate up and downsampling formula as:
\begin{itemize}
    \setlength{\itemindent}{-0.2cm}
    \item Upsampling:
    \begin{equation}
        \hskip-10pt\resizebox{0.8\columnwidth}{!}{$
        X_{LN \times MN} \approx \begin{bmatrix}
            \sqrt{LM} X_{N \times N} & \textbf{0}_{N \times (MN - N)}\\
            \textbf{0}_{(LN - N) \times N}  & \textbf{0}_{(LN-N) \times (MN-N)}
        \end{bmatrix}
        $}
    \end{equation}
    
    \item Downsampling:
    \begin{equation}
        X_{N\times N} \approx \frac{1}{\sqrt{LM}} X_{LN \times MN} \text{[0\,:\,$N$, 0\,:\,$N$]}
    \end{equation}
\end{itemize}

\section{Fourier transform's rotational property}
\label{apdx:fourierrotate}
The proof of the Fourier transform's rotational property is as follows. Define $g(\textbf{x})$ as a function of $x$ where $\textbf{x} \in \mathbb{R}^d$. The Fourier transform of $g$ is:
\begin{equation}
    \mathcal{F}[g(\textbf{x})] = G(\textbf{X}) = \int g(\textbf{x})e^{-j2\pi \textbf{x}^T\textbf{X}}\,d\textbf{x} 
\end{equation}

\noindent We can describe the rotated version of $\textbf{x}$ as $\textbf{u} = \textbf{Ax}$ where $\textbf{A}$ is a rotation matrix in which
\begin{align}
    \textbf{A}^{T} = \textbf{A}^{-1}\\
    \textbf{x} = \textbf{A}^{-1}\textbf{u} = \textbf{A}^{T}\textbf{u}
\end{align}

\noindent Define the rotated version of $g$ as $h$ where $g(\textbf{Ax}) = h(\textbf{x})$. Then, the Fourier transform of $g(\textbf{Ax})$ becomes:
\begin{align}
    \mathcal{F}[g(\textbf{Ax})] &= \mathcal{F}[h(\textbf{x})] = \int h(\textbf{x})e^{-j2\pi\textbf{x}^T\textbf{X}}\,d\textbf{x}\\
    &= \int g(\textbf{Ax})e^{-j2\pi\textbf{x}^T\textbf{X}}\,d\textbf{x}\\
    &= \int g(\textbf{u})e^{-j2\pi(\textbf{A}^T\textbf{u})^T\textbf{X}}\,d\textbf{u}\\
    &\hskip0.7cm {\footnotesize(\because d\textbf{u} = |\det(\textbf{A})|d\textbf{x},\;\; |\det(\textbf{A})| = 1)}\\
    &= \int g(\textbf{u})e^{-j2\pi\textbf{u}^T \textbf{AX}}\,d\textbf{u}\\
    \mathcal{F}[g(\textbf{Ax})] &= \int g(\textbf{u})e^{-j2\pi\textbf{u}^T \textbf{AX}}\,d\textbf{u} = G(\textbf{AX})
\end{align}

\noindent Thus, the Fourier transform of the rotated $g(\textbf{x})$ is equal to rotating the Fourier transform $G(\textbf{X})$.

\section{DCT to DFT sub-block conversion}
\label{apdx:fourierdecomp}
If we define $\omega = \exp(-j2\pi / N)$ then the $N \times N$ 1-D DFT bases matrix $W^{N \times N}$ is given as:
\begin{equation}
    \resizebox{0.87\linewidth}{!}{$
    W^{N \times N} = \frac{1}{\sqrt{N}}\begin{bmatrix}
        1 & 1 & 1 & \cdots & 1\\
        1 & \omega & \omega^2 & \cdots & \omega^{N-1}\\
        1 & \omega^2 & \omega^4 & \cdots & \omega^{2(N-1)}\\
        \vdots & \vdots & \vdots & \ddots & \vdots\\
        1 & \omega^{N-1} & \omega^{2(N-1)} & \cdots & \omega^{(N-1)(N-1)}
    \end{bmatrix}
    $}
\end{equation}
\noindent Setting $D_{N \times M}$ as the DFT coefficient block of size $N \times M$, the conversion formula becomes:
\begin{equation}
    \label{eq:subblockDFTcomb}
    \resizebox{0.87\linewidth}{!}{$
    D_{LN\times MN} = \hat{A}_{L,N}\begin{bmatrix}
        X_{N\times N}^{0,0} & \cdots & X_{N\times N}^{0,M-1} \\
        \vdots & \ddots & \vdots \\
        X_{N\times N}^{L-1,0} & \cdots & X_{N\times N}^{L-1,M-1}
    \end{bmatrix}\hat{A}^T_{M,N}
    $}
\end{equation}
The corresponding decomposition is then:
\begin{equation}
    \label{eq:subblockDFTdecomp}
    \resizebox{0.87\linewidth}{!}{$
    \begin{bmatrix}
        X_{N\times N}^{0,0} & \cdots & X_{N\times N}^{0,M-1} \\
        \vdots & \ddots & \vdots \\
        X_{N\times N}^{L-1,0} & \cdots & X_{N\times N}^{L-1,M-1}
    \end{bmatrix} = \hat{A}^{-1}_{L,N} D_{LN\times MN} \hat{A}^{-1^T}_{M,N}
    $}
\end{equation}
Where $\hat{A}$ denotes the DCT to DFT conversion matrix. This can be calculated by following the same process from \cref{eq:conversion_matrix} with replacing $B_{large}$ as $W$ of appropriate size.

\section{Resize strategy for DCT}
\label{apdx:resizestrategy}
While it is possible to do an arbitrary resize of $\frac{P}{Q} \times \frac{R}{S}$ by first upsampling $P \times R$ times and downsampling by $Q \times S$, it is preferable to avoid it due to the compute cost of an additional resize. Therefore, we utilize a different strategy. During random resized crop, we fuse the cropping and resize together in a way that the crop size is limited to the factors of a resize target. For example, if we are resizing to \,$\! 28 \times 28 \times 8 \times 8$, then the height and width of the crop window are selected from the set: $\{1, 2, 4, 7, 14, 28, 56, ...\}$. 
This way, we can reduce computation as upsampling or downsampling is limited to an integer multiple.
This strategy has been used throughout our experiments.

\section{Comparison with Pillow-SIMD}

Pillow-SIMD \cite{pillowsimd} is a highly optimized version of Pillow \cite{pillow}. It allows faster resizing using CPU SIMD instructions. The main bottleneck of our pipeline is resizing. If we reduce resizing by pre-resizing the data to $256 \times 256$, then our method shows 9.1\% and 29.2\% faster training and evaluation versus Pillow-SIMD as shown in \cref{tab:simd}. We believe our method could be further sped up given analogous optimization efforts. \vspace{0.2cm}

\begin{table}[h]
    \centering
    \renewcommand\cellset{\renewcommand\arraystretch{0.8}
    \setlength\extrarowheight{1pt}}
    \resizebox{\linewidth}{!}{
    \begin{tabular}{ccc| c c c c c c}
        Method & \makecell[b]{Img. size} & Model & \makecell[b]{Train Data} & \makecell[b]{Fwd/Bwd} & \makecell[b]{Train} & \makecell[b]{Eval Data} & \makecell[b]{Fwd} & \makecell[b]{Eval}\\
        \hline
        {SIMD} & $512^2$ & ViT-Ti & 1043.3 & 835.2 & 724.7 & 1743.9 & 2832.5 & 1630.8\\
        Ours & $512^2$ & ViT-Ti & \makecell[c]{816.2\\\small(-21.8\%)} & \makecell[c]{857.2\\\small(+2.6\%)} & \makecell[c]{687.6\\\small(-5.1\%)} & \makecell[c]{775.3\\\small(-55.5\%)} & \makecell[c]{2847.5\\\small(+0.5\%)} & \makecell[c]{752.3\\\small(-53.9\%)}\\
        \hline
        {SIMD} & $256^2$ & ViT-Ti & 1176.4 & 839.6 & 722.4 & 2232.0 & 2832.6 & 1893.9\\
        Ours & $256^2$ & ViT-Ti & \makecell[c]{1530.9\\\small(+30.1\%)} & \makecell[c]{854.2\\\small(+1.7\%)} & \makecell[c]{788.0\\\small\textbf{(+9.1\%)}} & \makecell[c]{3070.6\\\small(+37.6\%)} & \makecell[c]{2842.9\\\small(+0.4\%)} & \makecell[c]{2447.1\\\small\textbf{(+29.2\%)}}
    \end{tabular}}
    \caption{Comparison with Pillow-SIMD. Our pipeline shows faster training and evaluation if we reduce resizing.}
    \label{tab:simd}
\end{table}

\section{Smaller patch sizes} 
\label{apdx:smallerpatch}
SwinV2 \cite{liu2021swinv2} uses a patch size of 4. $8\times8$ JPEG DCT blocks can be adapted to support this by decomposing them into sixteen $2 \times 2$ blocks using sub-block conversion (Sec 5.2.). Then, we can use any of the embedding strategies discussed in Sec. 4. In general, for a desired patch size $p$, we need to decompose the DCT blocks to be at most $\frac{p}{2} \times \frac{p}{2}$.

\section{Training settings}
\label{apdx:trainingsettings}
    The hyperparameter settings and augmentation subset we used for training are reported in \Cref{tab:hyperparam,tab:augmentsubset}. RGB models used the recipe given in \cite{betterbaselinevit}, including the SwinV2 models \cite{liu2021swinv2}, for a fair comparison. \vspace{0.1cm}
\begin{table}[h]
    \centering
    \resizebox{0.9\columnwidth}{!}{
    \begin{tabular}{c|ccccc}
        Model & \makecell[b]{Learning\\Rate} & \makecell[b]{Weight\\Decay} & \makecell[b]{RandAug.\\Magnitude} & \makecell[b]{Input\\Size} & \makecell[c]{Epochs} \\
        \hline
        ViT-Ti & 1e-3 & 1e-4 & 10 & 224$^2$ & 300\\
        ViT-S & 1e-3 & 1e-4 & 10 & 224$^2$ & 90\\
        SwinV2-T (RGB) & 1e-3 & 1e-4 & 10 & 256$^2$ & 300\\
        \hline
        JPEG-Ti & 3e-3 & 1e-4 & 3 & 224$^2$ & 300\\
        JPEG-S & 3e-3 & 3e-4 & 3 & 224$^2$ & 90\\
        SwinV2-T (DCT) & 1e-3 & 1e-4 & 3 & 256$^2$ & 300
    \end{tabular}
    }
    \caption{Hyperparameter settings of the trained models.}
    \label{tab:hyperparam}
\end{table}\vspace{0.1cm}

\begin{table}[h]
    \centering
    \resizebox{0.9\columnwidth}{!}{
    \begin{tabular}{c|c}
        Models & Subset \\
        \hline
        JPEG-Ti & \makecell[c]{\fontsize{8}{10}\selectfont Brightness, Contrast, Color, AutoContrast, \\\fontsize{8}{10}\selectfont AutoSaturation, MidfreqAug, Posterize, SolarizeAdd, \\\fontsize{8}{10}\selectfont Grayscale, ChromaDrop, Translate, Cutout, Rotate90}\\
        \hline
        \makecell[c]{JPEG-S\\SwinV2-T (DCT)} & \makecell[c]{\fontsize{8}{10}\selectfont Brightness, Contrast, Color, AutoContrast, \\\fontsize{8}{10}\selectfont AutoSaturation, MidfreqAug, Sharpness, Posterize, \\\fontsize{8}{10}\selectfont Grayscale, ChromaDrop, Translate, Cutout, Rotate90}\\
        \hline
        RGB models\cite{betterbaselinevit} & \makecell[c]{\fontsize{8}{10}\selectfont Brightness, Contrast, Equalize, Color, \\\fontsize{8}{10}\selectfont AutoContrast, Sharpness, Invert, Posterize, Solarize, \\\fontsize{8}{10}\selectfont SolarizeAdd, Translate, Cutout, Rotate, Shear}
    \end{tabular}
    }
    \caption{Augmentation subset of RandAugment for the models.}
    \label{tab:augmentsubset}
\end{table}

\section{Measurement process}
\label{apdx:measurement}
\textbf{Latency} measurements to decode and augment follow \cref{alg:latency_decode,alg:latency_augment}.
\textbf{Data Load}ing throughputs for both train and evaluation is measured using \cref{alg:throughput_dataload}.
\textbf{Model Fwd/Bwd} measures the throughput of model forward and backward pass using \cref{alg:throughput_modelfbp}.
\textbf{Model Fwd} measures the throughput of the model forward pass using \cref{alg:throughput_modelfwd}.
\textbf{Train Pipeline} and \textbf{Eval Pipeline} throughput is measured using \cref{alg:throughput_trainpipe,alg:throughput_evalpipe}. \vspace{0.3cm}

\newcommand{\algsize}{\footnotesize} 
\begin{algorithm}[H]
    \caption{Decoding latency measurement}
    \label{alg:latency_decode}
    \algsize
    \begin{algorithmic}
        \State{latency $\gets$ 0}
        \For{$i=0..N$}
            \State{start\_time $\gets$ \texttt{time}()}
            \State{data $\gets$ \texttt{decode}(Filename)}
            \State{end\_time $\gets$ \texttt{time}()}
            \State{latency $\gets$ latency + (end\_time $-$ start\_time)}
        \EndFor
        \State{\textbf{return} latency$/N$}
    \end{algorithmic}
\end{algorithm}

\begin{algorithm}[H]
    \caption{Augment latency measurement}
    \label{alg:latency_augment}
    \algsize
    \begin{algorithmic}
        \State{latency $\gets$ 0}
        \For{$i=0..N$}
            \State{data $\gets$ \texttt{decode}(Filename)}
            \State{start\_time $\gets$ \texttt{time}()}
            \State{data $\gets$ \texttt{augment}(data)}
            \State{end\_time $\gets$ \texttt{time}()}
            \State{latency $\gets$ latency + (end\_time $-$ start\_time)}
        \EndFor
        \State{\textbf{return} latency$/N$}
    \end{algorithmic}
\end{algorithm}

\begin{algorithm}[H]
    \caption{Data Loading throughput measurement}
    \label{alg:throughput_dataload}
    \algsize
    \begin{algorithmic}
        \State{start\_time $\gets$ \texttt{time}()}
        \For{$i=0..N$}
            \State{data, label $\gets$ \texttt{to\_gpu}(\texttt{next}(data\_loader))}
        \EndFor
        \State{end\_time $\gets$ \texttt{time}()}
        \State{latency $\gets$ end\_time $-$ start\_time}
        \State{\textbf{return} $(N \cdot \texttt{len}(\text{data}))/$latency}
    \end{algorithmic}
\end{algorithm}

\begin{algorithm}[H]
    \caption{Model Fwd/Bwd throughput measurement}
    \label{alg:throughput_modelfbp}
    \algsize
    \begin{algorithmic}
        \State{dummy\_data, dummy\_label $\gets$ \texttt{to\_gpu}(\texttt{random}(data\_shape))}
        \State{start\_time $\gets$ \texttt{time}()}
        \For{$i=0..N$}
            \State{data, label $\gets$ \texttt{copy}(dummy\_data, dummy\_label)}
            \State{\texttt{mixup}(data, label)}
            \State{output $\gets$ \texttt{model}(data)}
            \State{loss $\gets$ \texttt{criterion}(output, label)}
            \State{\texttt{backward}(loss)}
            \State{\texttt{step}(optimizer)}
        \EndFor
        \State{end\_time $\gets$ \texttt{time}()}
        \State{latency $\gets$ end\_time $-$ start\_time}
        \State{\textbf{return} $(N \cdot \texttt{len}(\text{dummy\_data}))/$latency}
    \end{algorithmic}
\end{algorithm}

\begin{algorithm}[H]
    \caption{Model Fwd throughput measurement}
    \label{alg:throughput_modelfwd}
    \algsize
    \begin{algorithmic}
        \State{dummy\_data, dummy\_label $\gets$ \texttt{to\_gpu}(\texttt{random}(data\_shape))}
        \State{start\_time $\gets$ \texttt{time}()}
        \For{$i=0..N$}
            \State{output $\gets$ \texttt{model}(dummy\_data)}
            \State{loss $\gets$ \texttt{criterion}(output, dummy\_label)}
        \EndFor
        \State{end\_time $\gets$ \texttt{time}()}
        \State{latency $\gets$ end\_time $-$ start\_time}
        \State{\textbf{return} $(N \cdot \texttt{len}(\text{dummy\_data}))/$latency}
    \end{algorithmic}
\end{algorithm}

\begin{algorithm}[H]
    \caption{Train Pipeline throughput measurement}
    \label{alg:throughput_trainpipe}
    \algsize
    \begin{algorithmic}
        \State{start\_time $\gets$ \texttt{time}()}
        \For{$i=0..N$}
            \State{data, label $\gets$ \texttt{to\_gpu}(\texttt{next}(data\_loader))}
            \State{\texttt{mixup}(data, label)}
            \State{output $\gets$ \texttt{model}(data)}
            \State{loss $\gets$ \texttt{criterion}(output, label)}
            \State{\texttt{backward}(loss)}
            \State{\texttt{step}(optimizer)}
        \EndFor
        \State{end\_time $\gets$ \texttt{time}()}
        \State{latency $\gets$ end\_time $-$ start\_time}
        \State{\textbf{return} $(N \cdot \texttt{len}(\text{data}))/$latency}
    \end{algorithmic}
\end{algorithm}

\begin{algorithm}[H]
    \caption{Eval Pipeline throughput measurement}
    \label{alg:throughput_evalpipe}
    \algsize
    \begin{algorithmic}
        \State{start\_time $\gets$ \texttt{time}()}
        \For{$i=0..N$}
            \State{data, label $\gets$ \texttt{to\_gpu}(\texttt{next}(data\_loader))}
            \State{output $\gets$ \texttt{model}(data)}
            \State{loss $\gets$ \texttt{criterion}(output, label)}
        \EndFor
        \State{end\_time $\gets$ \texttt{time}()}
        \State{latency $\gets$ end\_time $-$ start\_time}
        \State{\textbf{return} $(N \cdot \texttt{len}(\text{data}))/$latency}
    \end{algorithmic}
\end{algorithm}

\end{appendices}

\end{document}